\documentclass[10pt,twocolumn,letterpaper]{article}

\usepackage[pagenumbers]{cvpr} 

\usepackage{graphicx}
\usepackage{amsmath}
\usepackage{amssymb}
\usepackage{booktabs}

\usepackage[accsupp]{axessibility}  
\usepackage[table]{xcolor}  
\usepackage{arydshln}
\usepackage{amsmath}
\usepackage{float}
\usepackage{multirow}
\usepackage{pifont}    
\usepackage{listings}  
\usepackage{xcolor}    
\def\ours{\textbf{Ours}}
\def\adain{\textbf{Ours}-\texttt{AdaIN}}

\newcommand{\h}{15mm}
\newcommand{\hsrc}{0.5mm}
\newcommand{\himg}{-0.7mm}
\newcommand{\hdataset}{0.6mm}
\newcommand{\htask}{10mm}
\newcommand{\hlong}{80mm}

\newcommand{\figframework}{
\begin{figure*}[t]
    \vspace{-2mm}
    \centering
    \includegraphics[width=0.90\linewidth]{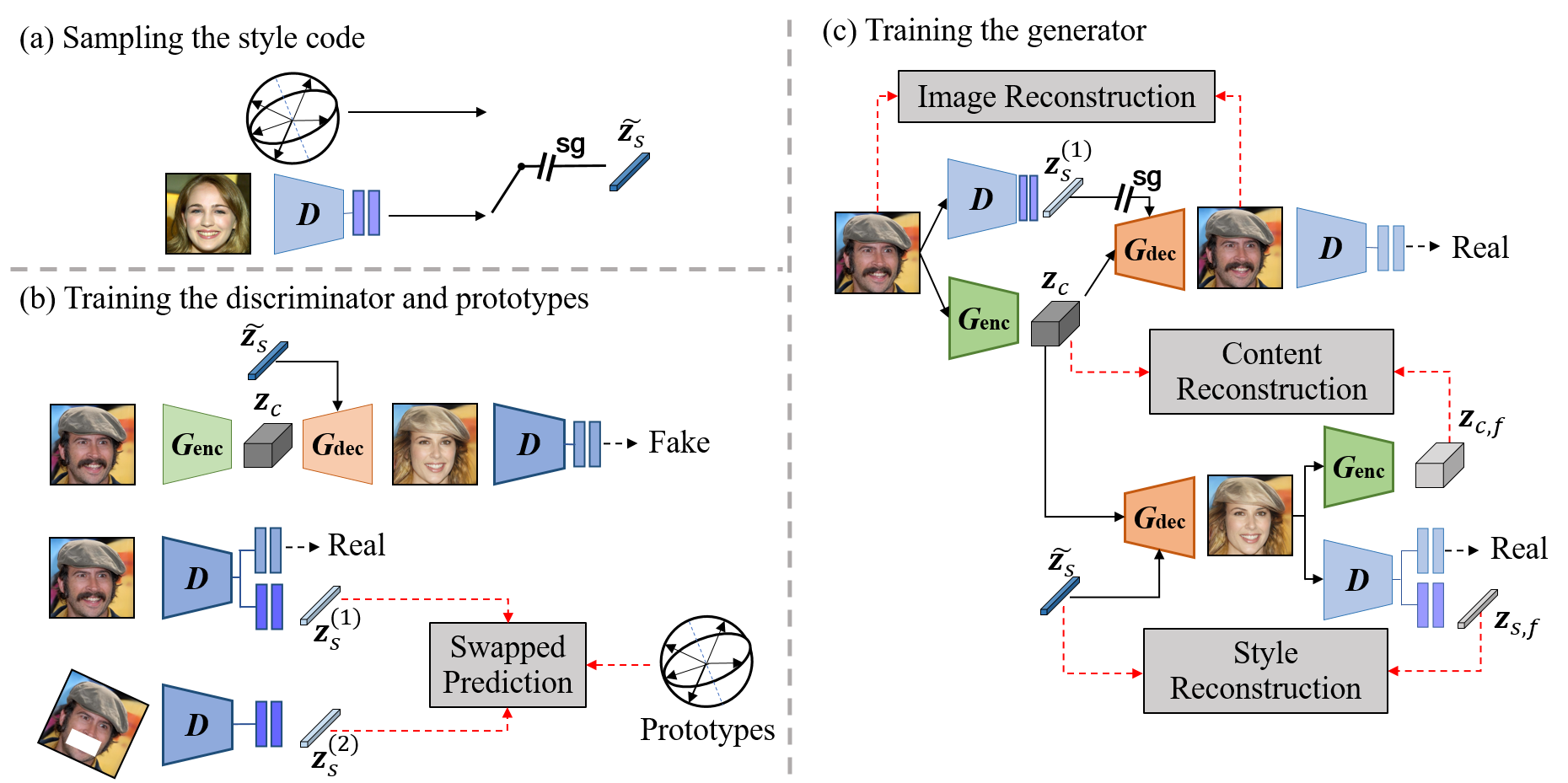}
    \vspace{-1mm}
    \caption{Framework overview. (a) The style code is sampled from the learned prototypes or dataset. (b) The discriminator not only learns to distinguish between real and fake images but also learns the style space via the swapped prediction loss. (c) The generator is enforced to utilize the style code via the style reconstruction and to preserve the input content via the content reconstruction.}
    \label{fig:framework}
\end{figure*}
}

\newcommand{\figlatent}{
\renewcommand{\h}{14mm}
\renewcommand{\hsrc}{0.1mm}
\renewcommand{\himg}{-0.8mm}
\renewcommand{\hlong}{85mm}  
\begin{figure*}[t]
    \vspace{-2mm}
    \centering
    \footnotesize{
        \makebox[\h][c]{\textbf{Source}}\hspace{\hsrc}
        \makebox[\h][c]{}\hspace{\himg}
        \makebox[\h][c]{}\hspace{\himg}
        \makebox[\h][c]{}\hspace{\himg}
        \makebox[\h][c]{}\hspace{\himg}
        \makebox[\h][c]{}\hspace{\hdataset}
        \makebox[\h][c]{\textbf{Source}}\hspace{\hsrc}
        \makebox[\h][c]{}\hspace{\himg}
        \makebox[\h][c]{}\hspace{\himg}
        \makebox[\h][c]{}\hspace{\himg}
        \makebox[\h][c]{}\hspace{\himg}
        \makebox[\h][c]{}
    }\\
    \includegraphics[width=\h]{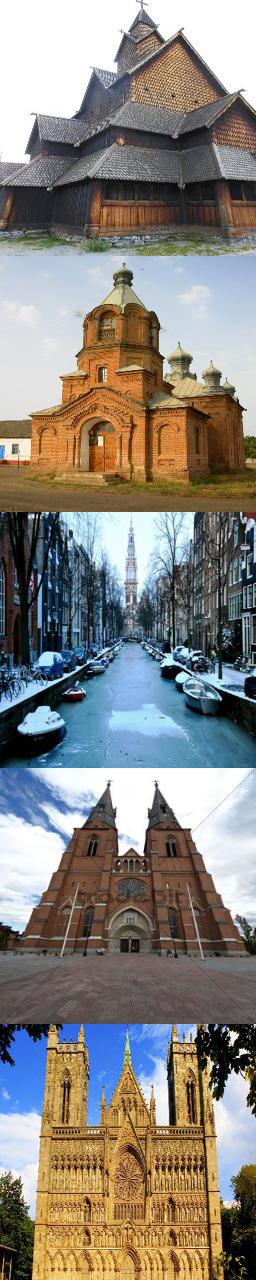}\hspace{\hsrc}
    \includegraphics[width=\h]{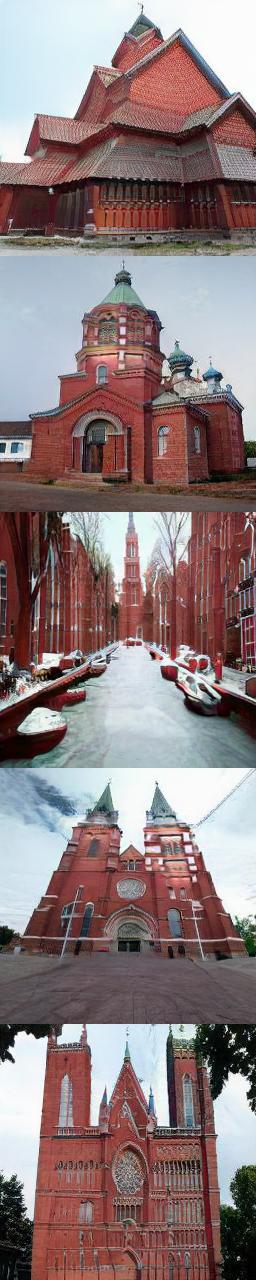}\hspace{\himg}
    \includegraphics[width=\h]{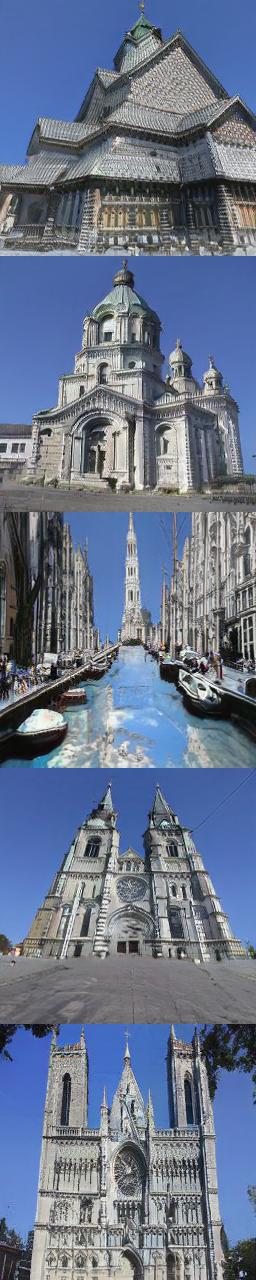}\hspace{\himg}
    \includegraphics[width=\h]{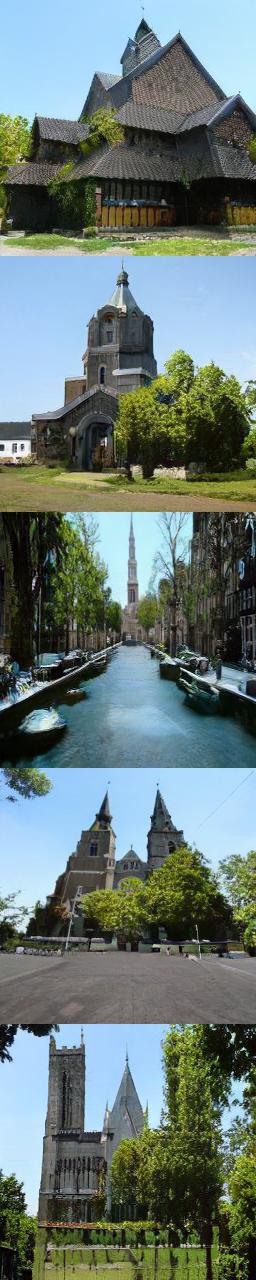}\hspace{\himg}
    \includegraphics[width=\h]{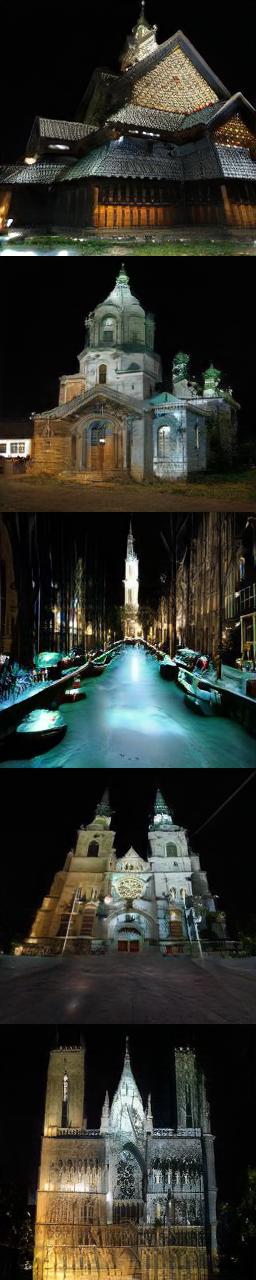}\hspace{\himg}
    \includegraphics[width=\h]{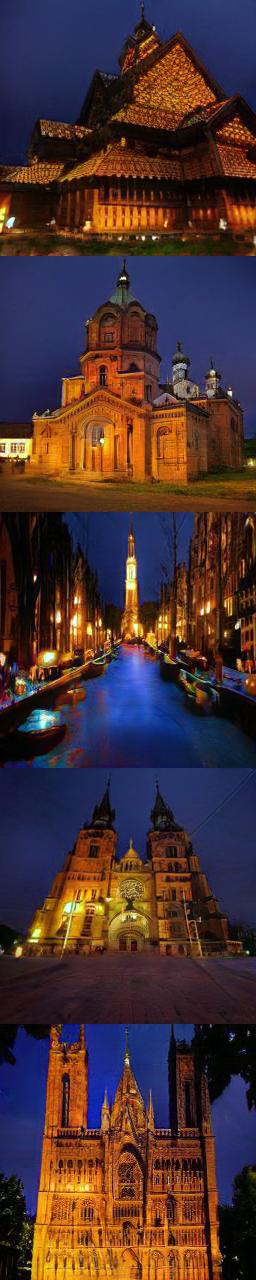}\hspace{\hdataset}
    \includegraphics[width=\h]{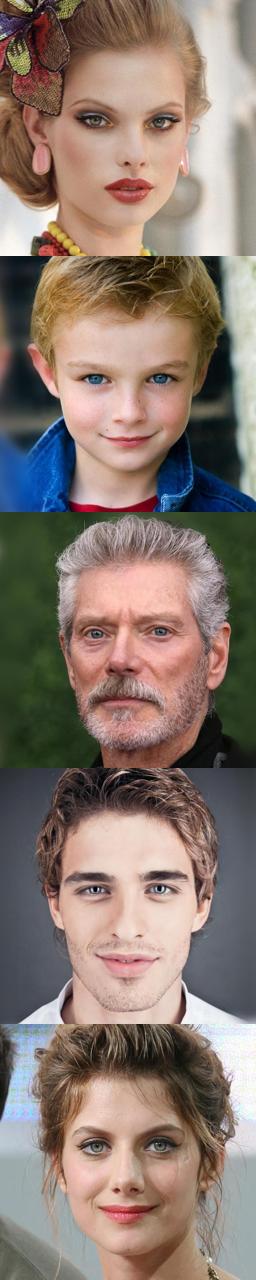}\hspace{\hsrc}
    \includegraphics[width=\h]{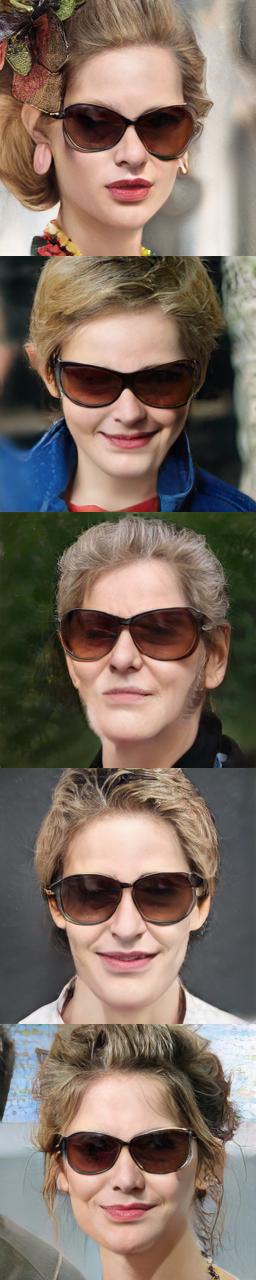}\hspace{\himg}
    \includegraphics[width=\h]{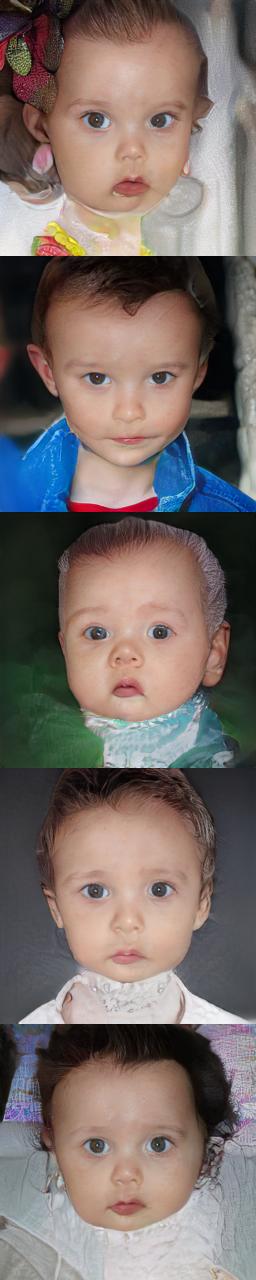}\hspace{\himg}
    \includegraphics[width=\h]{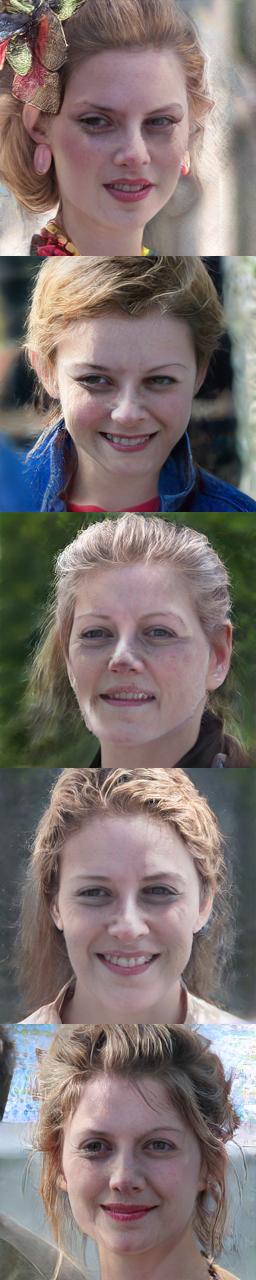}\hspace{\himg}
    \includegraphics[width=\h]{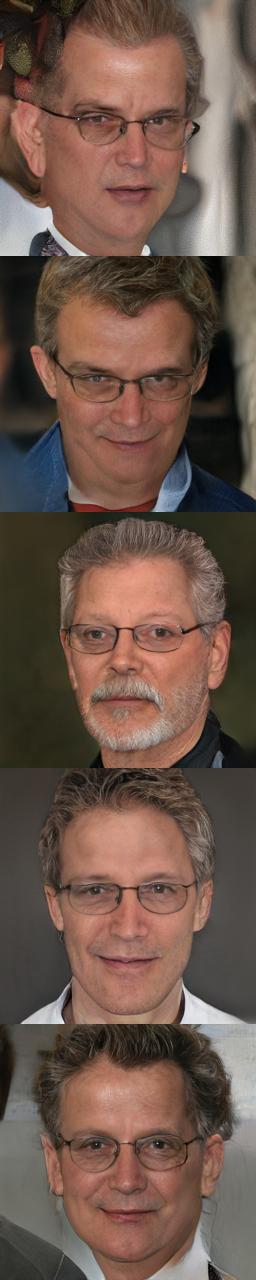}\hspace{\himg}
    \includegraphics[width=\h]{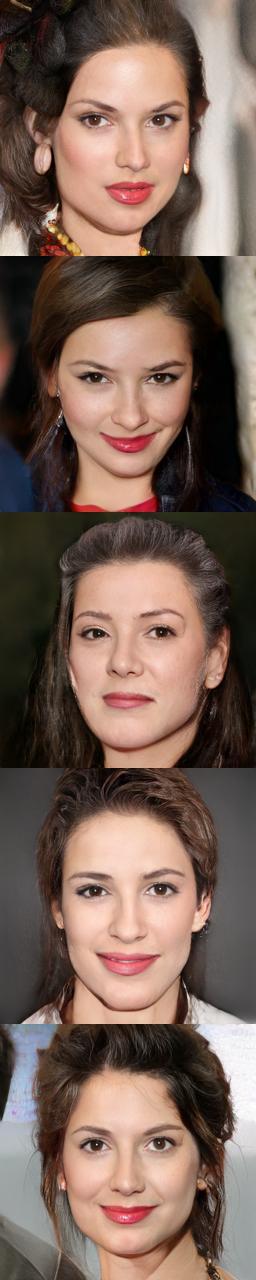}\vspace{0.1mm}
    \makebox[\hlong][c]{LSUN churches}\hspace{\hdataset}
    \makebox[\hlong][c]{FFHQ}
    \vspace{-1mm}
    \caption{Prototype-guided synthesis. Our model discovers various style prototypes from the dataset in an unsupervised manner. The style prototype consists of a combination of varioius attributes including (left) time, weather, season, and texture; and (right) age, gender, and accessories. Each row shows the result of manipulating the leftmost image with learned prototypes.}
    \vspace{-1mm}
    \label{fig:latent}
\end{figure*}
}

\newcommand{\figref}{
\renewcommand{\h}{17mm}
\renewcommand{\hsrc}{0.1mm}
\renewcommand{\himg}{-0.8mm}
\begin{figure*}[t]
    \vspace{-2mm}
    \centering
    \footnotesize{
        \makebox[\h][c]{\textbf{Source}}\hspace{\himg}
        \makebox[\h][c]{\textbf{Reference}}\hspace{\hsrc}
        \makebox[\h][c]{\ours}\hspace{\himg}
        \makebox[\h][c]{\adain}\hspace{\himg}
        \makebox[\h][c]{StarGAN v2}\hspace{\himg}
        \makebox[\h][c]{Liu \etal \cite{liu2021smoothing}}\hspace{\himg}
        \makebox[\h][c]{CLUIT}\hspace{\himg}
        \makebox[\h][c]{SwapAE}\hspace{\himg}
        \makebox[\h][c]{StyleMapGAN}
    }\\
    \includegraphics[width=\h]{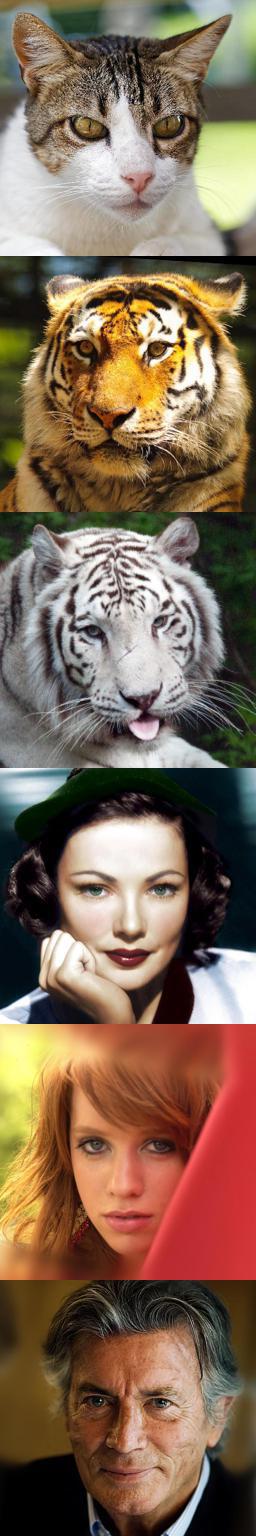}\hspace{\himg}
    \includegraphics[width=\h]{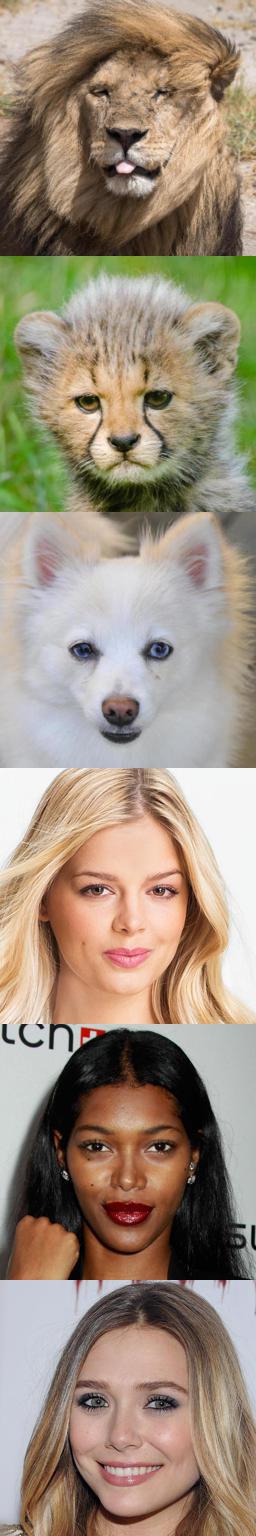}\hspace{\hsrc}
    \includegraphics[width=\h]{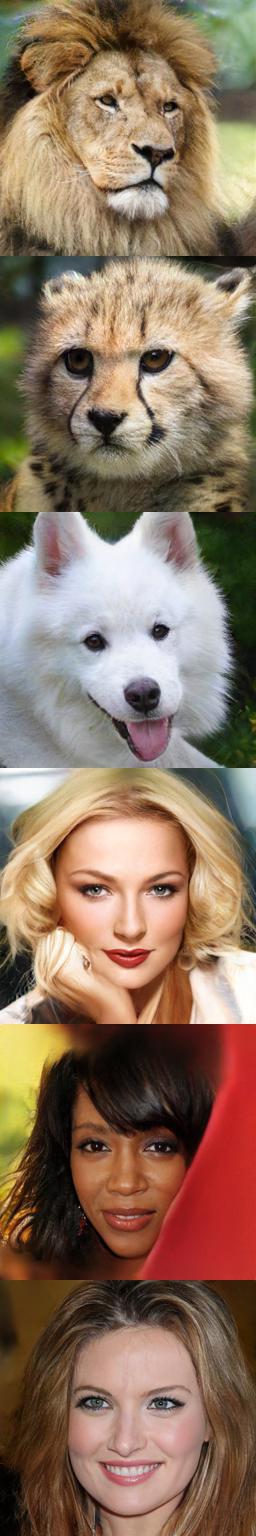}\hspace{\himg}
    \includegraphics[width=\h]{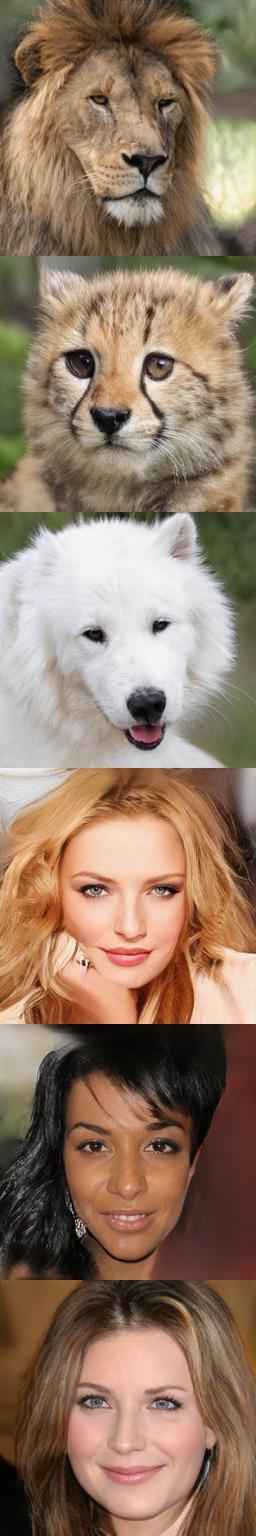}\hspace{\himg}
    \includegraphics[width=\h]{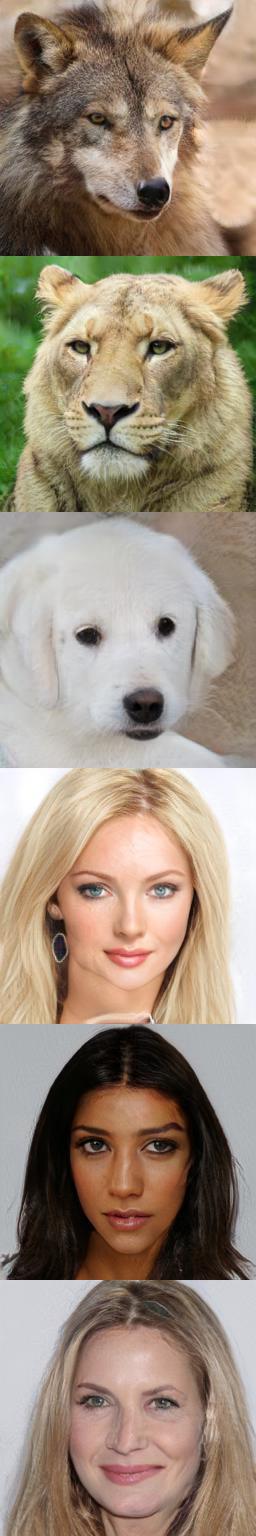}\hspace{\himg}
    \includegraphics[width=\h]{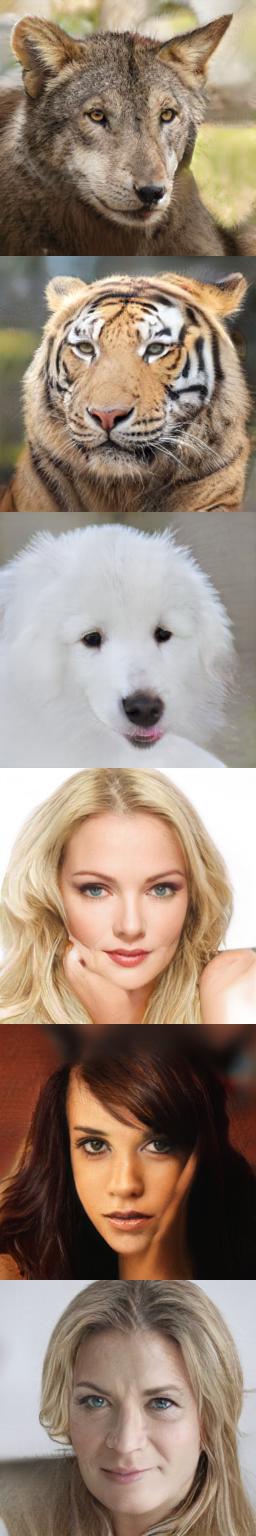}\hspace{\himg}
    \includegraphics[width=\h]{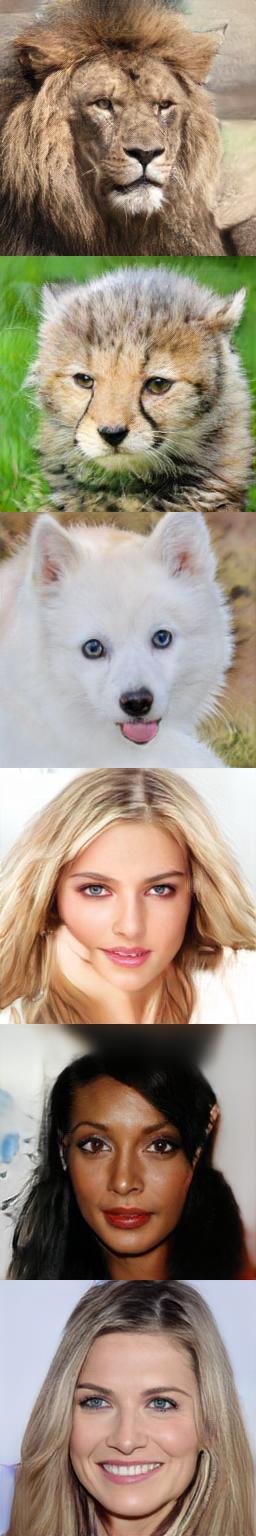}\hspace{\himg}
    \includegraphics[width=\h]{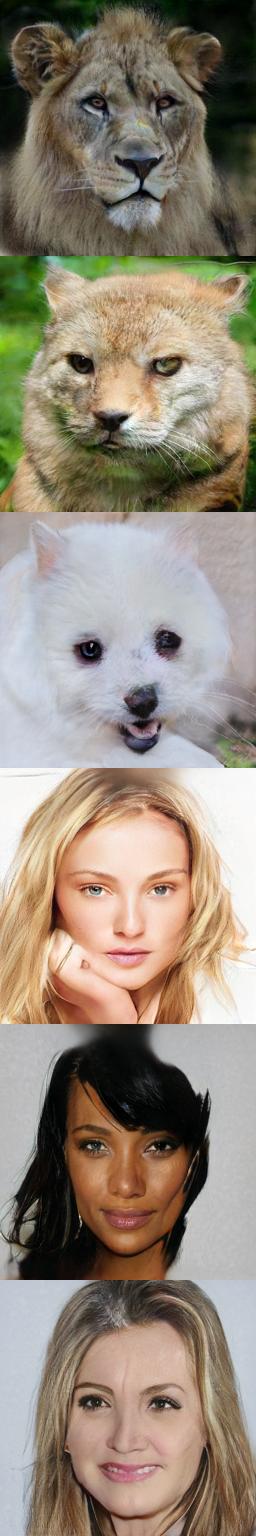}\hspace{\himg}
    \includegraphics[width=\h]{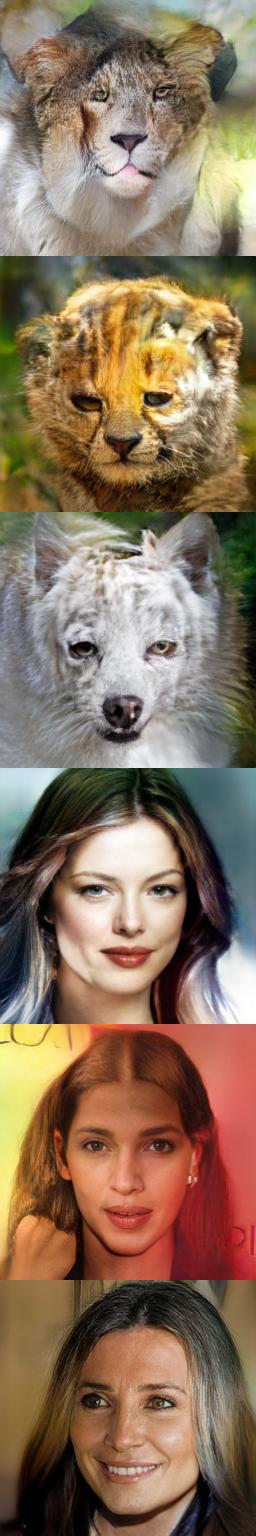}\vspace{-2mm}
    \caption{
        Qualitative comparison of reference-guided image synthesis on AFHQ (top three rows) and CelebA-HQ (bottom three rows).
    }
    \vspace{-1mm}
    \label{fig:reference}
\end{figure*}
}

\newcommand{\figcontrol}{
\renewcommand{\h}{18mm}
\renewcommand{\hsrc}{0.3mm}
\renewcommand{\himg}{-0.7mm}
\renewcommand{\htask}{5mm}
\renewcommand{\hlong}{70.9mm}
\begin{figure*}[t]
    \vspace{-2mm}
    \centering
    \footnotesize{
        \makebox[\h][c]{\textbf{Source}}\hspace{\hsrc}
        \makebox[\h][c]{\textbf{Reconstruction}}\hspace{\hsrc}
        \makebox[\h][c]{}\hspace{\himg}
        \makebox[\h][c]{}\hspace{\himg}
        \makebox[\h][c]{}\hspace{\himg}
        \makebox[\h][c]{}\hspace{\himg}
        \makebox[\h][c]{}\hspace{\himg}
        \makebox[\h][c]{}
    }\\
    \includegraphics[width=\h]{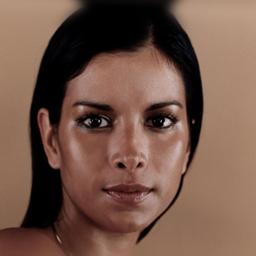}\hspace{\hsrc}
    \includegraphics[width=\h]{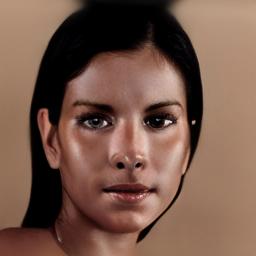}\hspace{\hsrc}
    \includegraphics[width=\h]{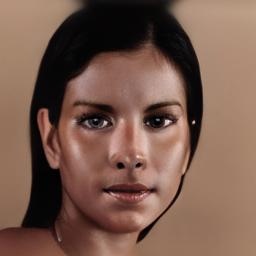}\hspace{\himg}
    \includegraphics[width=\h]{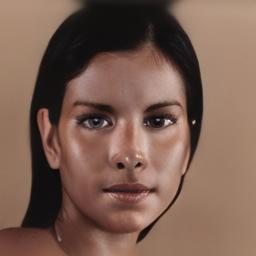}\hspace{\himg}
    \includegraphics[width=\h]{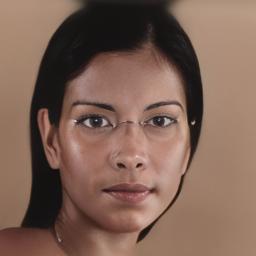}\hspace{\himg}
    \includegraphics[width=\h]{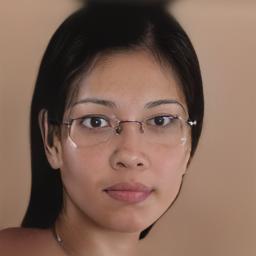}\hspace{\himg}
    \includegraphics[width=\h]{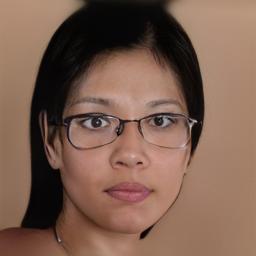}\hspace{\himg}
    \includegraphics[width=\h]{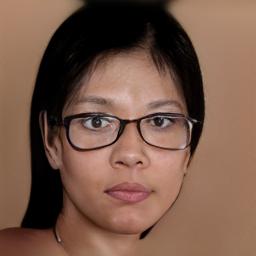}\\
    \includegraphics[width=\h]{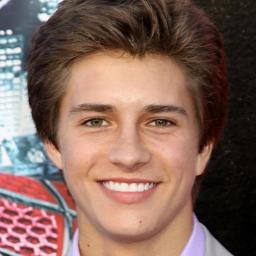}\hspace{\hsrc}
    \includegraphics[width=\h]{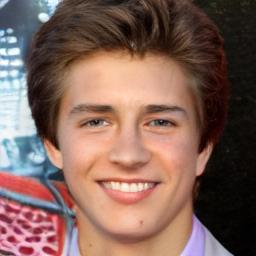}\hspace{\hsrc}
    \includegraphics[width=\h]{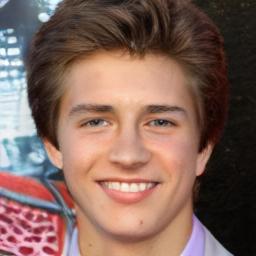}\hspace{\himg}
    \includegraphics[width=\h]{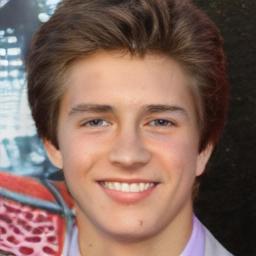}\hspace{\himg}
    \includegraphics[width=\h]{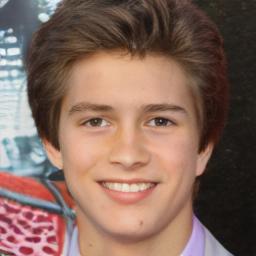}\hspace{\himg}
    \includegraphics[width=\h]{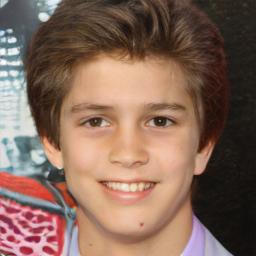}\hspace{\himg}
    \includegraphics[width=\h]{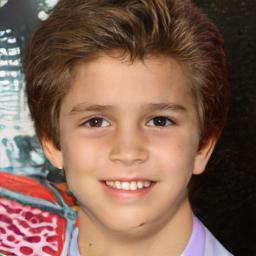}\hspace{\himg}
    \includegraphics[width=\h]{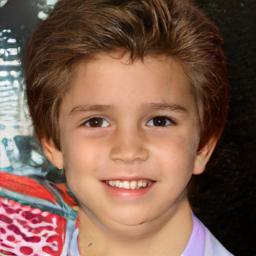}\\
    \includegraphics[width=\h]{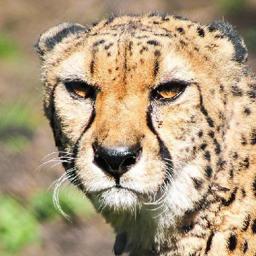}\hspace{\hsrc}
    \includegraphics[width=\h]{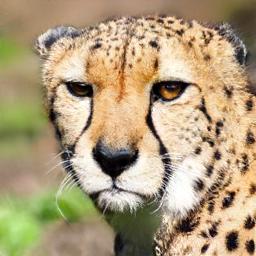}\hspace{\hsrc}
    \includegraphics[width=\h]{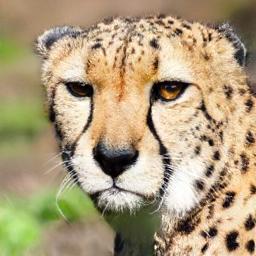}\hspace{\himg}
    \includegraphics[width=\h]{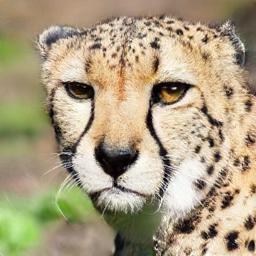}\hspace{\himg}
    \includegraphics[width=\h]{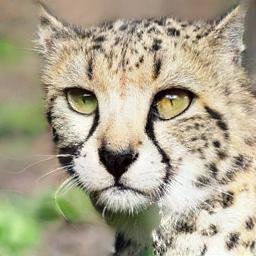}\hspace{\himg}
    \includegraphics[width=\h]{figures/04/a/1_4.jpg}\hspace{\himg}
    \includegraphics[width=\h]{figures/04/a/1_4.jpg}\hspace{\himg}
    \includegraphics[width=\h]{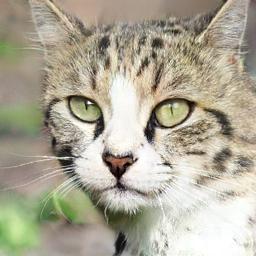}
    \makebox[141.1mm][c]{\footnotesize{(a) Reconstruction and style interpolation results on FFHQ, and AFHQ. The first two source images are from CelebA-HQ.}}\\\vspace{0.5mm}
    \footnotesize{
        \makebox[\h][c]{\textbf{Target}}\hspace{\himg}
        \makebox[\h][c]{\textbf{Original}}\hspace{\hsrc}
        \makebox[\h][c]{\textbf{Ours}}\hspace{\himg}
        \makebox[\h][c]{StyleMapGAN}\hspace{\htask}
        \makebox[\h][c]{\textbf{Source}}\hspace{\himg}
        \makebox[\h][c]{\textbf{Reference 1}}\hspace{\himg}
        \makebox[\h][c]{\textbf{Reference 2}}\hspace{\hsrc}
        \makebox[\h][c]{\textbf{Output}}
    }\\
    \includegraphics[width=\h]{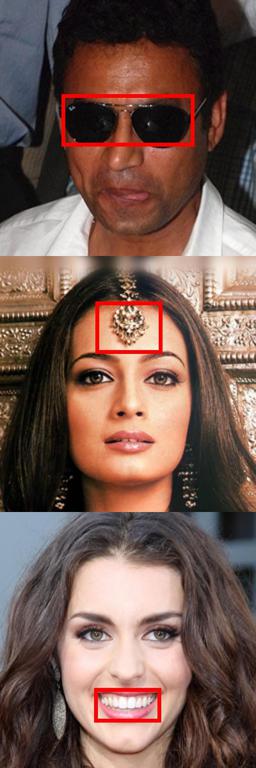}\hspace{\himg}
    \includegraphics[width=\h]{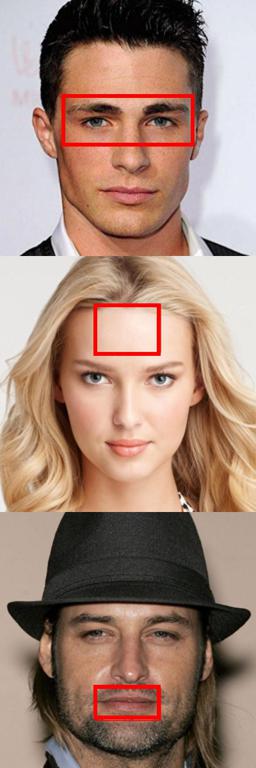}\hspace{\hsrc}
    \includegraphics[width=\h]{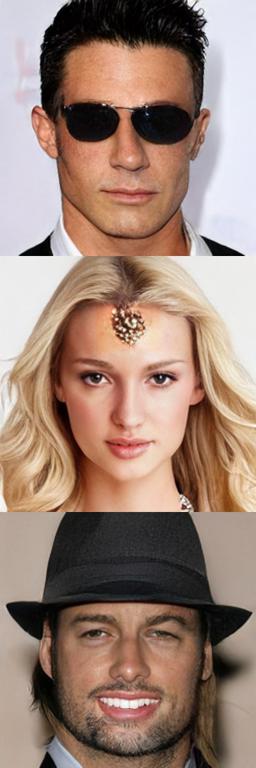}\hspace{\himg}
    \includegraphics[width=\h]{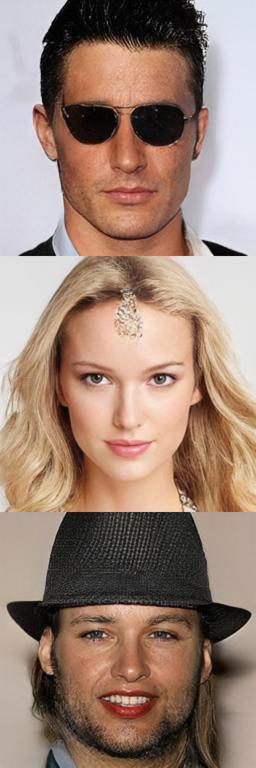}\hspace{\htask}
    \includegraphics[width=\h]{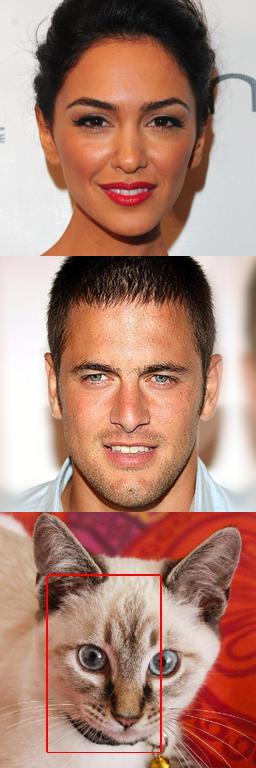}\hspace{\himg}
    \includegraphics[width=\h]{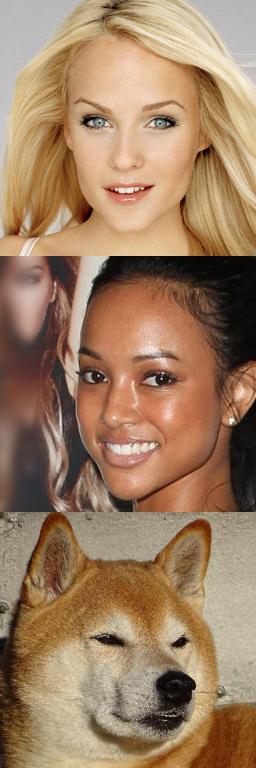}\hspace{\himg}
    \includegraphics[width=\h]{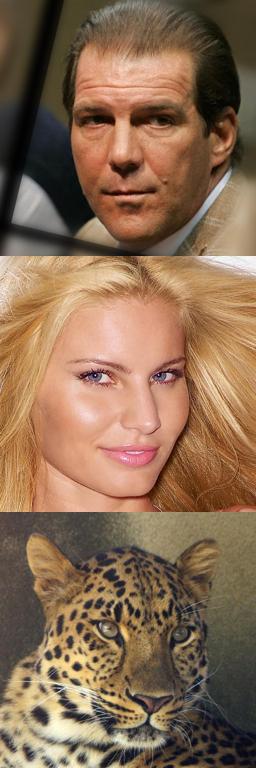}\hspace{\hsrc}
    \includegraphics[width=\h]{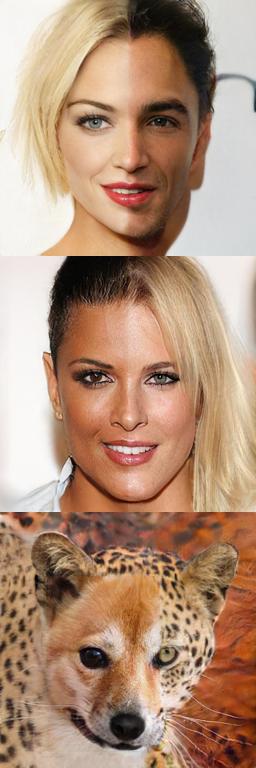}\\
    \makebox[\hlong][c]{\footnotesize{(b) Content transplantation comparison on CelebA-HQ}}\hspace{\htask}
    \makebox[\hlong][c]{\footnotesize{(c) Local image translation results on CelebA-HQ and AFHQ.}}\vspace{-1mm}
    \caption{Examples of various applications. The proposed method is capable of manipulating the style and content of an image in real-time.}
    \vspace{-2mm}
    \label{fig:control}
\end{figure*}
}

\newcommand{\figsearch}{
\renewcommand{\h}{14mm}
\renewcommand{\hsrc}{1.5mm}
\renewcommand{\hlong}{70mm}
\begin{figure*}[t]
    \vspace{-2mm}
    \centering
    \footnotesize{
        \makebox[\h][c]{\textbf{Query}}\hspace{\hsrc}
        \makebox[\hlong][c]{\textbf{w/o color distortion}}\hspace{\hsrc}
        \makebox[\hlong][c]{w/ color distortion}
    }\\
    \includegraphics[width=\h]{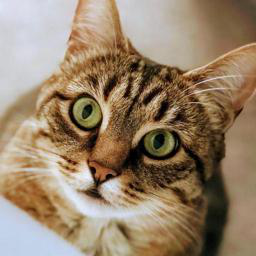}\hspace{\hsrc}
    \includegraphics[width=\h]{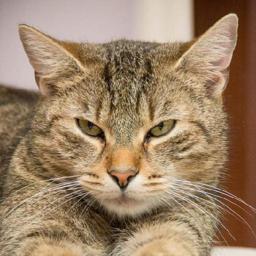}\hspace{\himg}
    \includegraphics[width=\h]{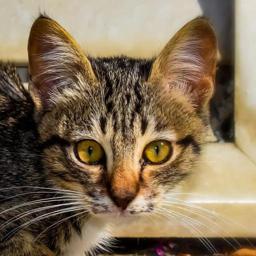}\hspace{\himg}
    \includegraphics[width=\h]{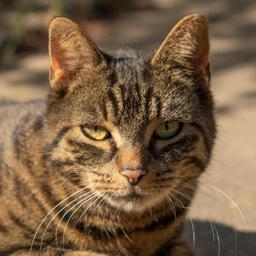}\hspace{\himg}
    \includegraphics[width=\h]{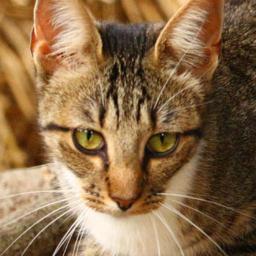}\hspace{\himg}
    \includegraphics[width=\h]{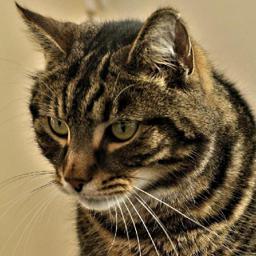}\hspace{\hsrc}
    \includegraphics[width=\h]{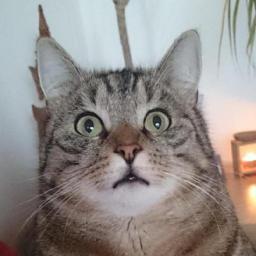}\hspace{\himg}
    \includegraphics[width=\h]{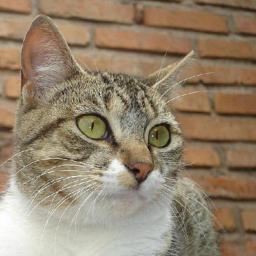}\hspace{\himg}
    \includegraphics[width=\h]{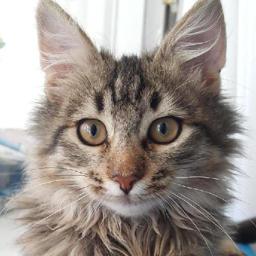}\hspace{\himg}
    \includegraphics[width=\h]{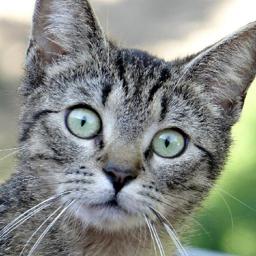}\hspace{\himg}
    \includegraphics[width=\h]{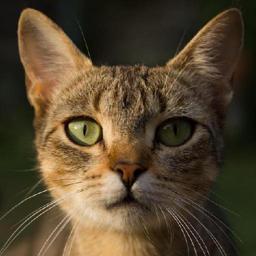}\\
    \footnotesize{
        \makebox[\h][c]{\textbf{Query}}\hspace{\hsrc}
        \makebox[\hlong][c]{\textbf{w/ cutout}}\hspace{\hsrc}
        \makebox[\hlong][c]{w/o cutout}
    }\\
    \includegraphics[width=\h]{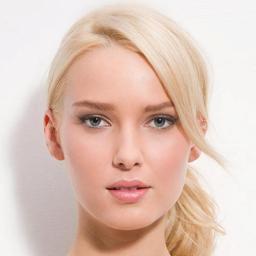}\hspace{\hsrc}
    \includegraphics[width=\h]{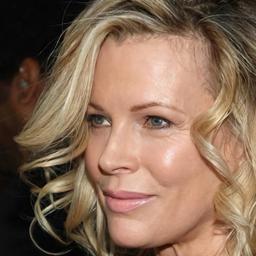}\hspace{\himg}
    \includegraphics[width=\h]{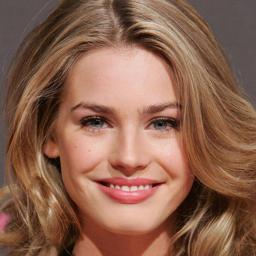}\hspace{\himg}
    \includegraphics[width=\h]{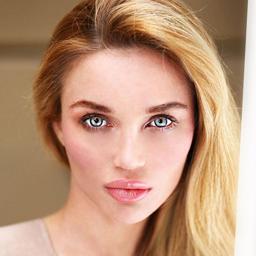}\hspace{\himg}
    \includegraphics[width=\h]{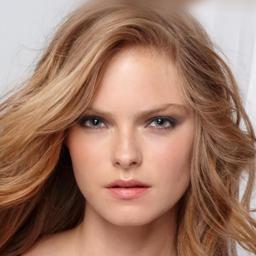}\hspace{\himg}
    \includegraphics[width=\h]{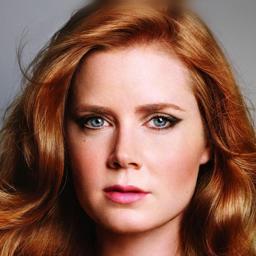}\hspace{\hsrc}
    \includegraphics[width=\h]{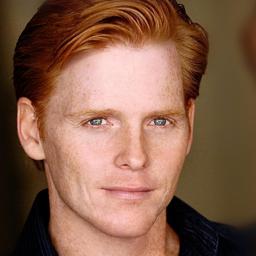}\hspace{\himg}
    \includegraphics[width=\h]{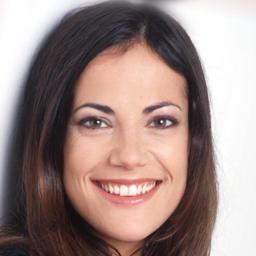}\hspace{\himg}
    \includegraphics[width=\h]{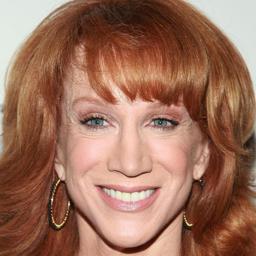}\hspace{\himg}
    \includegraphics[width=\h]{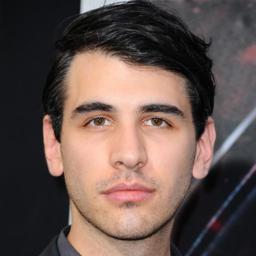}\hspace{\himg}
    \includegraphics[width=\h]{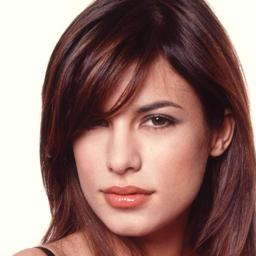}\\
    \vspace{-1mm}
    \caption{Similarity search results on the AFHQ and CelebA-HQ datasets. We projected the query and test set into the style space and performed a nearest neighbor search. We plot here the five most similar images in style space.}
    \label{fig:search}
    \vspace{-1mm}
\end{figure*}
}

\newcommand{\figablation}{
\renewcommand{\h}{18mm}
\renewcommand{\hsrc}{0.1mm}
\begin{figure}
    \vspace{-2mm}
    \centering
    \footnotesize{
        \makebox[\h][c]{\textbf{Source}}\hspace{\himg}
        \makebox[\h][c]{\textbf{Reference}}\hspace{\hsrc}
        \makebox[\h][c]{\textbf{w/o color dist.}}\hspace{\himg}
        \makebox[\h][c]{w/ color dist.}\hspace{\himg}
    }\\
    \includegraphics[width=\h]{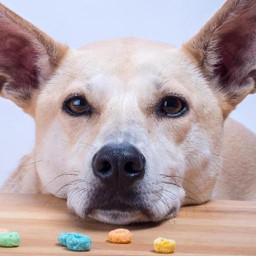}\hspace{\himg}
    \includegraphics[width=\h]{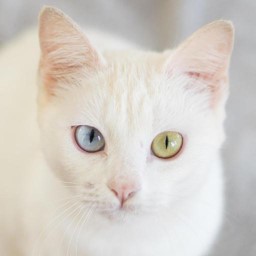}\hspace{\hsrc}
    \includegraphics[width=\h]{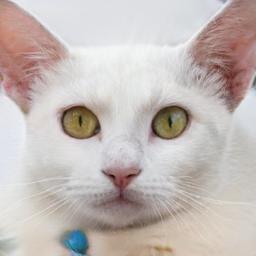}\hspace{\himg}
    \includegraphics[width=\h]{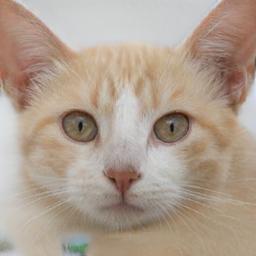}\hspace{\himg}\\\vspace{-0.5mm}
    \footnotesize{
        \makebox[\h][c]{\textbf{Source}}\hspace{\himg}
        \makebox[\h][c]{\textbf{Reference}}\hspace{\hsrc}
        \makebox[\h][c]{\textbf{w/ cut-out}}\hspace{\himg}
        \makebox[\h][c]{w/o cut-out}\hspace{\himg}
    }\\
    \includegraphics[width=\h]{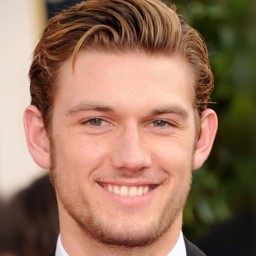}\hspace{\himg}
    \includegraphics[width=\h]{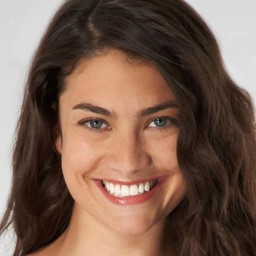}\hspace{\hsrc}
    \includegraphics[width=\h]{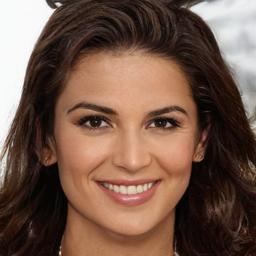}\hspace{\himg}
    \includegraphics[width=\h]{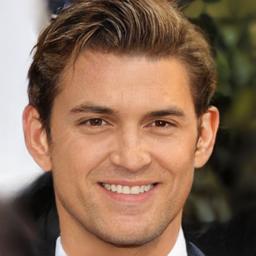}\hspace{\himg}
    \vspace{-1mm}
    \caption{Comparison of the results for various augmentations.}
    \vspace{-2mm}
    \label{fig:ablation}
\end{figure}
}

\newcommand{\figlerp}{
\begin{figure}
    \vspace{-2mm}
    \centering
    \includegraphics[width=0.98\linewidth]{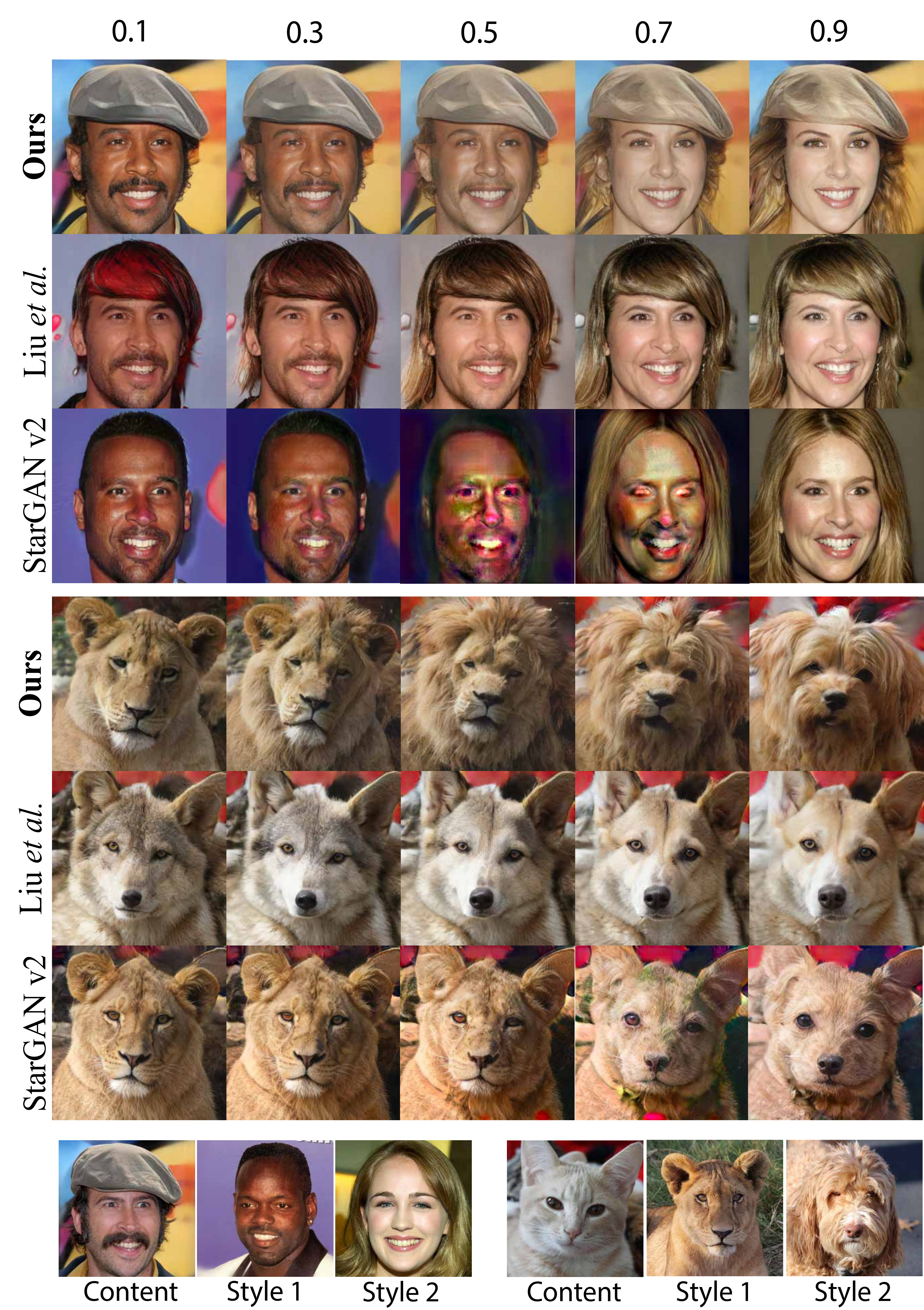}
    \vspace{-1mm}
    \caption{Qualitative comparison of the style interpolation. We sampled three images (one source and two references) from the dataset and synthesized images using the style code interpolated between the two style codes obtained from the two reference images.}
    \label{fig:lerp}
\end{figure}
}

\newcommand{\figsep}{
\begin{figure}
    \centering
    \includegraphics[width=0.95\linewidth]{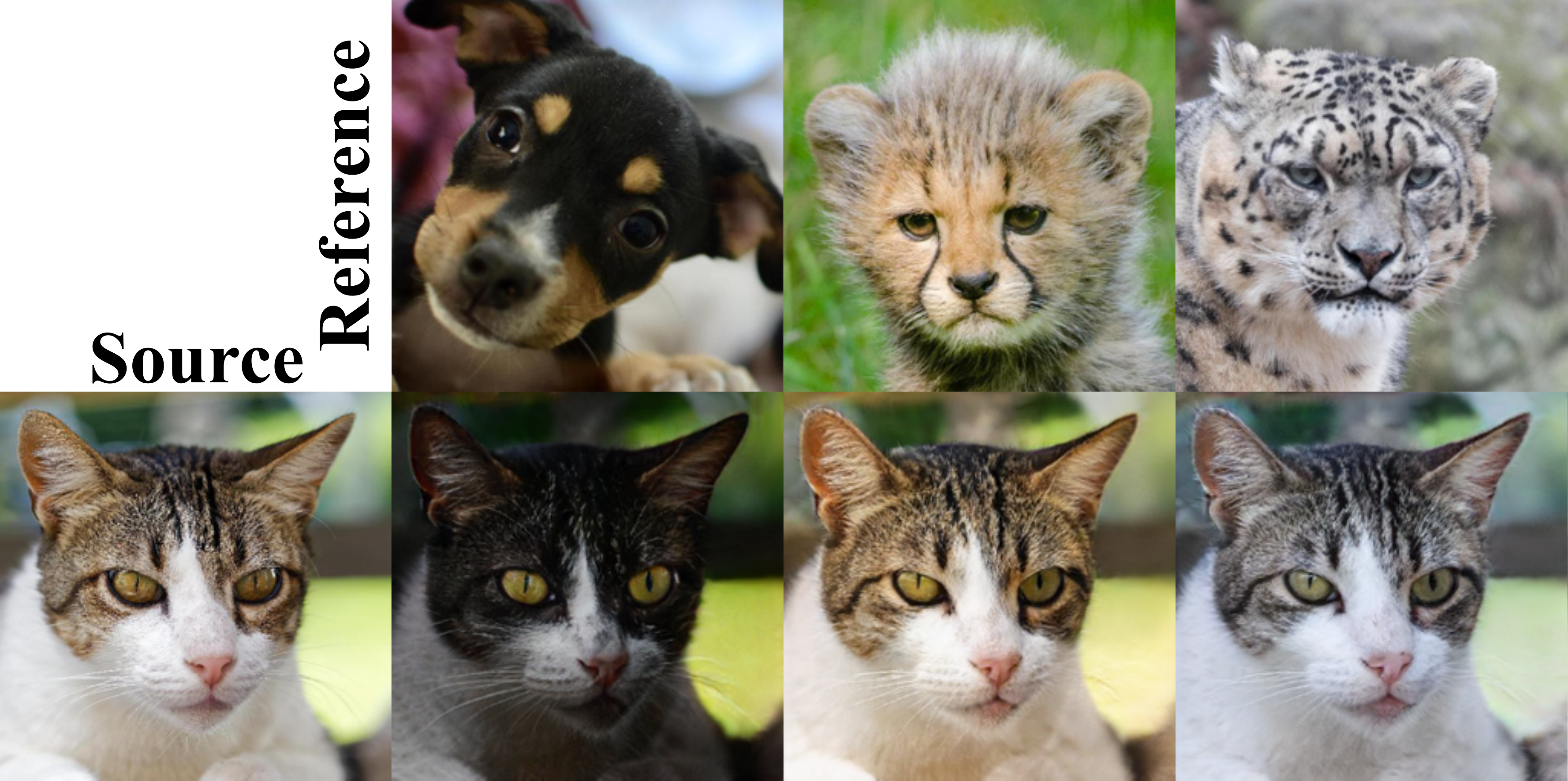}
    \vspace{-1mm}
    \caption{Qualitative results for the \texttt{separated} method.}
    \label{fig:sep}
\end{figure}
}

\newcommand{\figafhq}{
\renewcommand{\h}{33mm}
\renewcommand{\himg}{-0.9mm}
\renewcommand{\hsrc}{0.1mm}
\begin{figure*}[p!]
    \centering
    \includegraphics[width=\h]{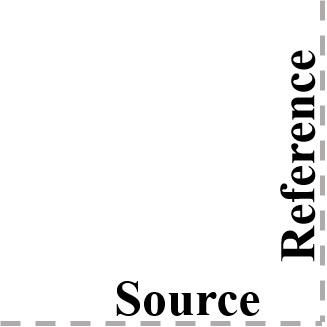}\hspace{\hsrc}
    \includegraphics[width=\h]{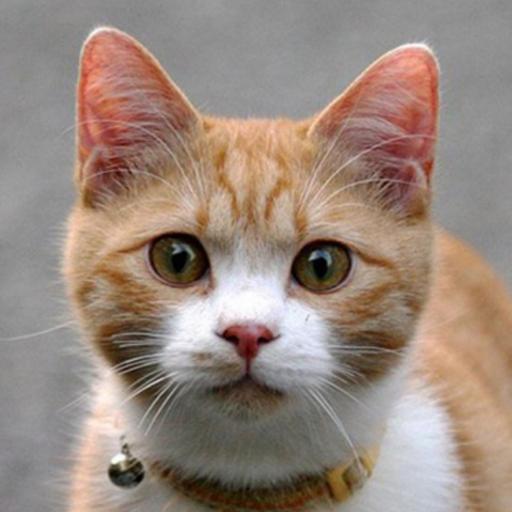}\hspace{\himg}
    \includegraphics[width=\h]{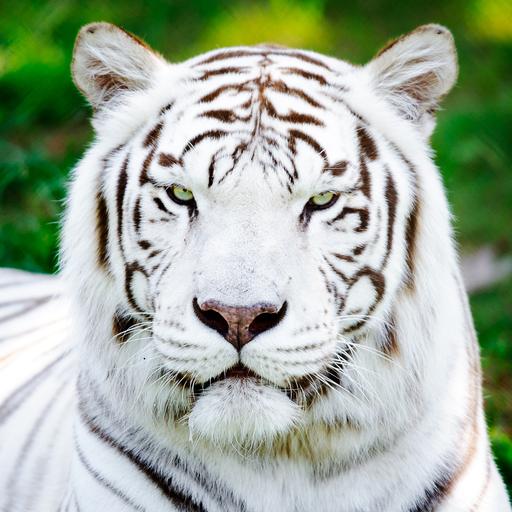}\hspace{\himg}
    \includegraphics[width=\h]{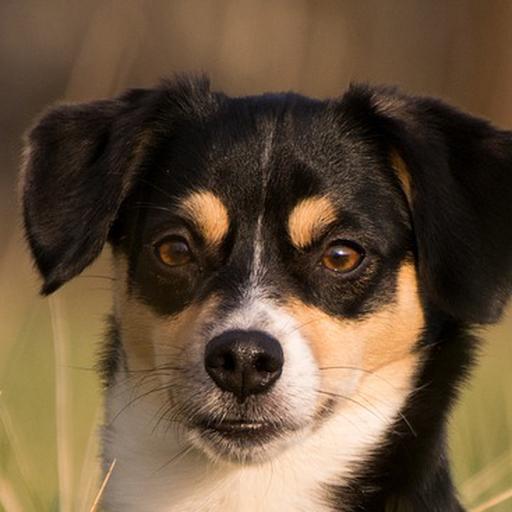}\hspace{\himg}
    \includegraphics[width=\h]{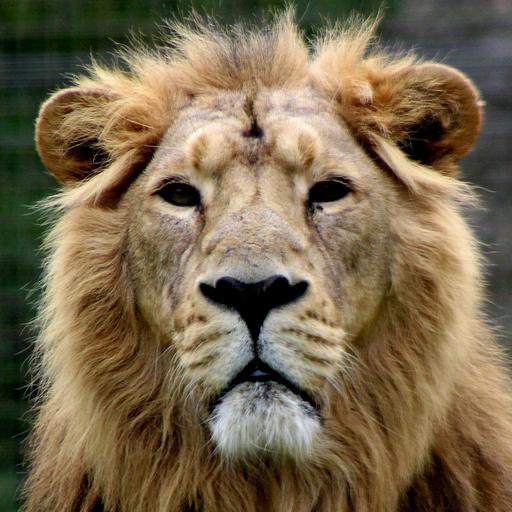}\\\vspace{1mm}
    \includegraphics[width=\h]{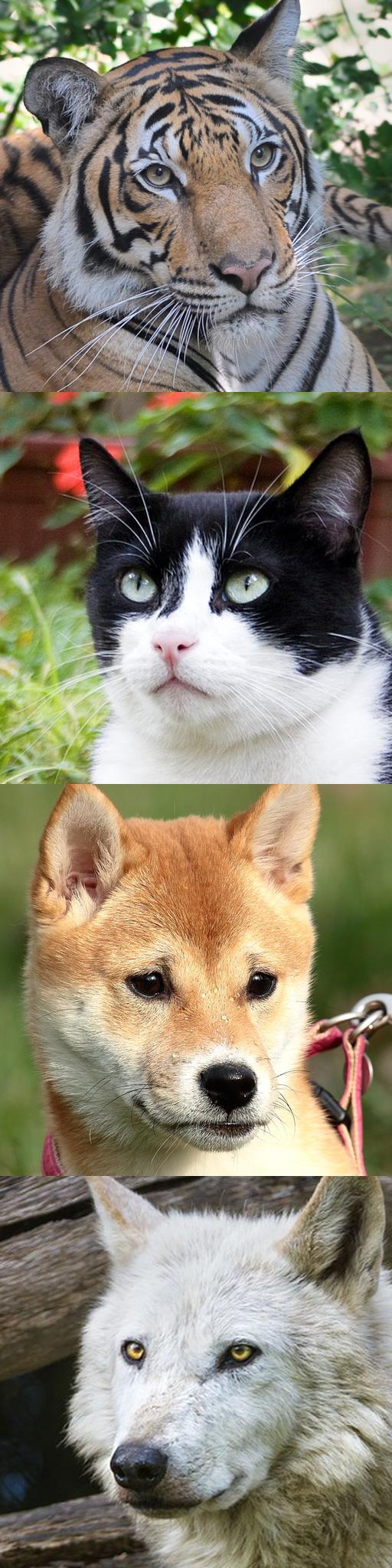}\hspace{\hsrc}
    \includegraphics[width=\h]{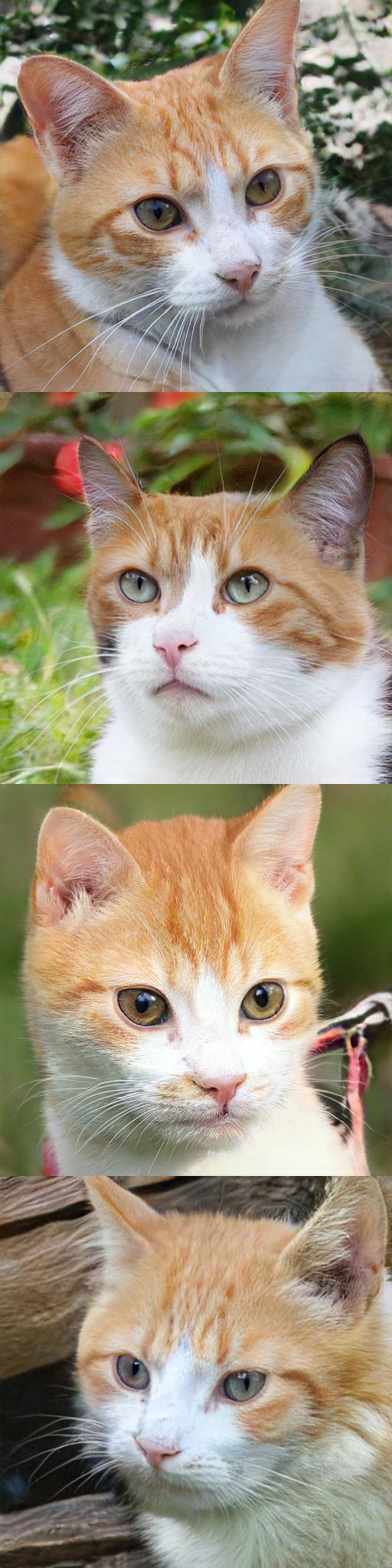}\hspace{\himg}
    \includegraphics[width=\h]{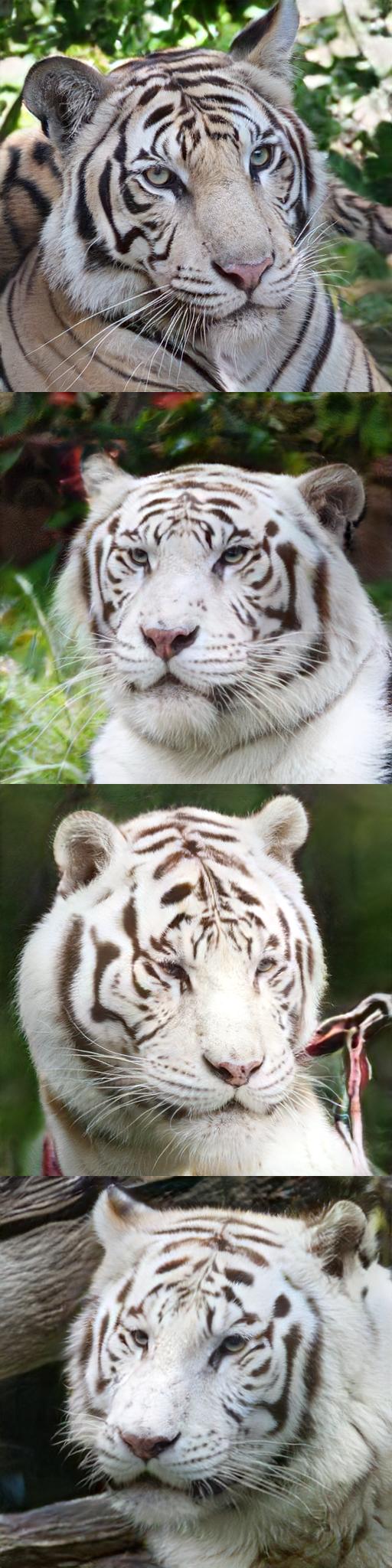}\hspace{\himg}
    \includegraphics[width=\h]{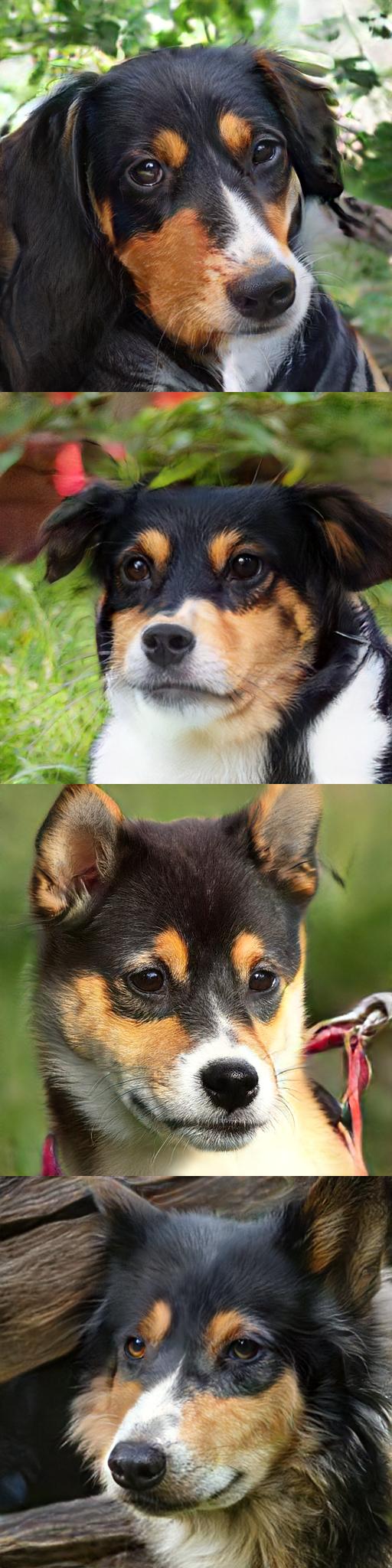}\hspace{\himg}
    \includegraphics[width=\h]{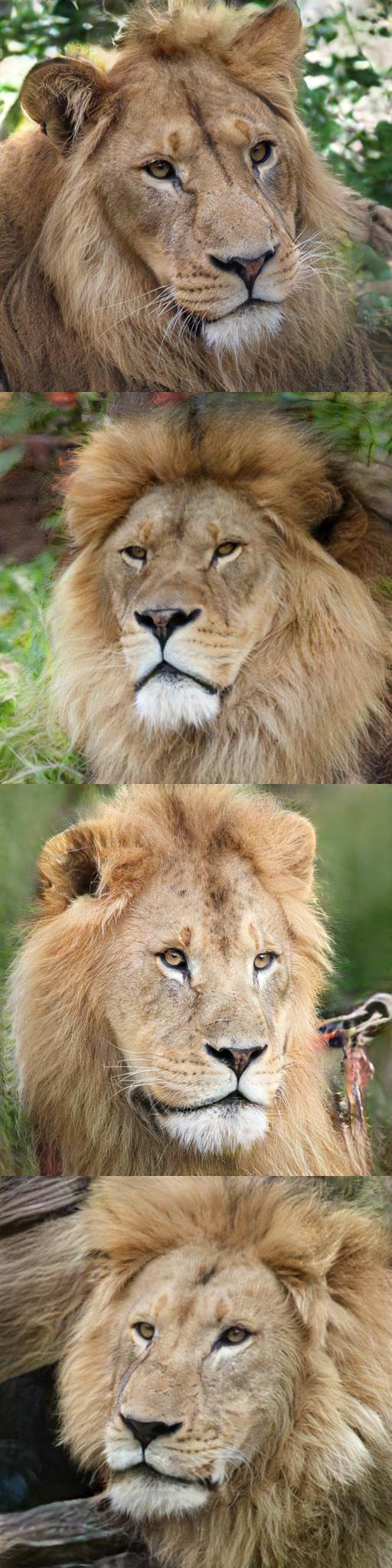}\\
    \caption{Reference-guided synthesis results on the AFHQ v2 dataset. The model was trained and tested at 512$\times$512 resolution.}
    \label{fig:afhqv2}
\end{figure*}
}

\newcommand{\figffhq}{
\renewcommand{\h}{33mm}
\renewcommand{\himg}{-0.9mm}
\renewcommand{\hsrc}{0.1mm}
\begin{figure*}[p!]
    \centering
    \includegraphics[width=\h]{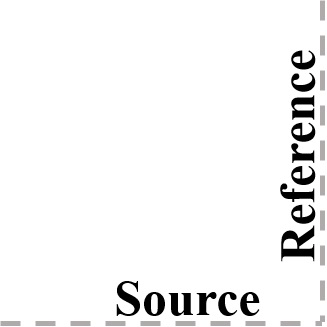}\hspace{\hsrc}
    \includegraphics[width=\h]{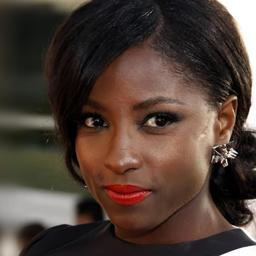}\hspace{\himg}
    \includegraphics[width=\h]{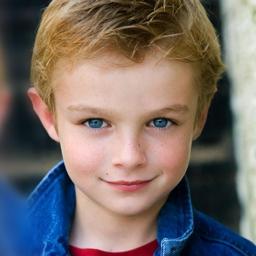}\hspace{\himg}
    \includegraphics[width=\h]{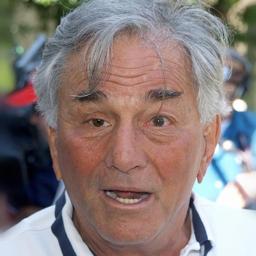}\hspace{\himg}
    \includegraphics[width=\h]{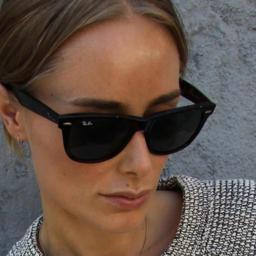}\\\vspace{1mm}
    \includegraphics[width=\h]{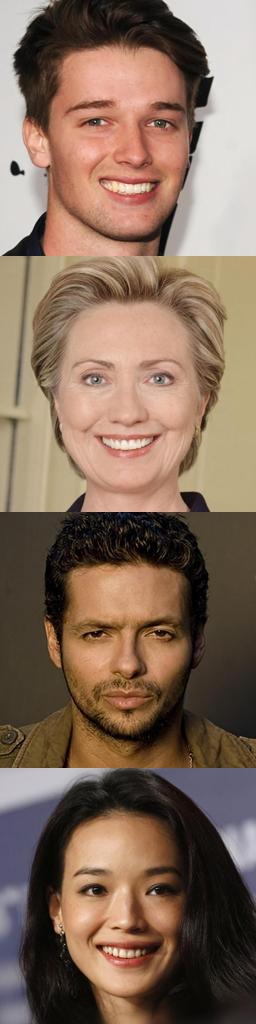}\hspace{\hsrc}
    \includegraphics[width=\h]{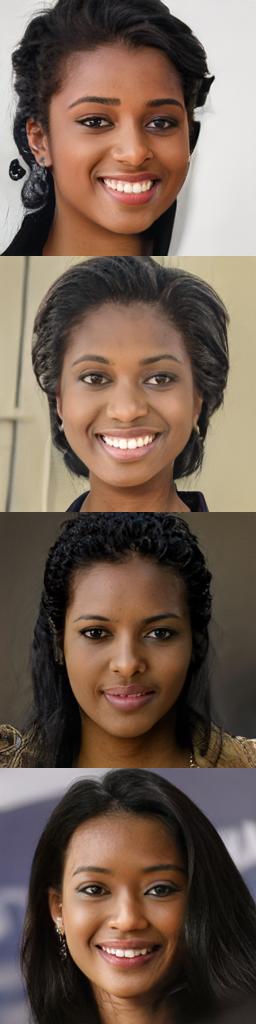}\hspace{\himg}
    \includegraphics[width=\h]{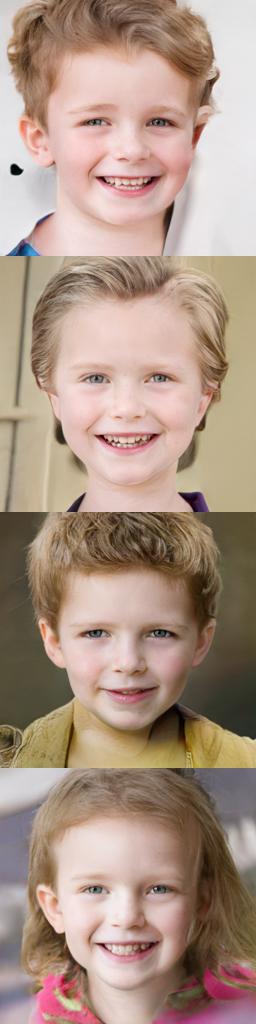}\hspace{\himg}
    \includegraphics[width=\h]{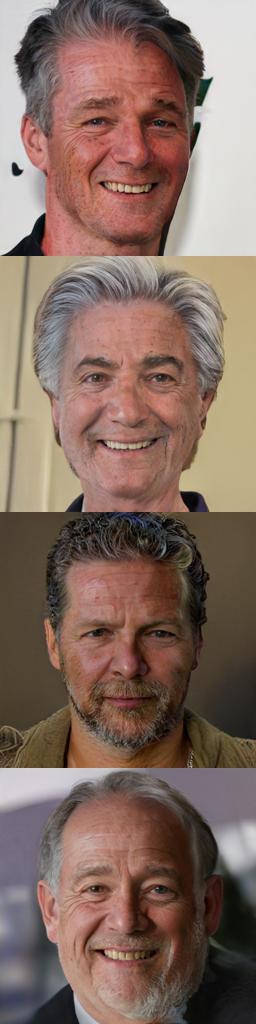}\hspace{\himg}
    \includegraphics[width=\h]{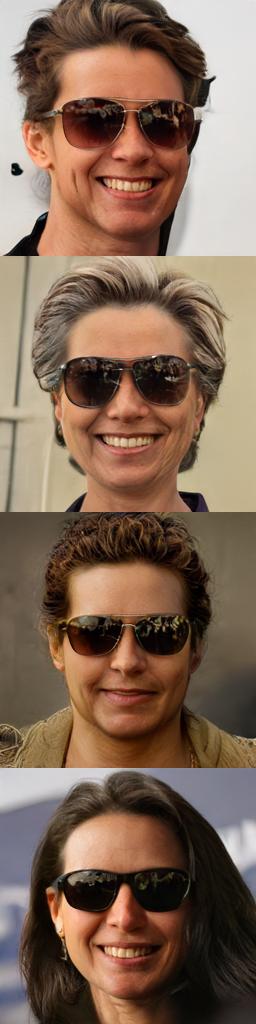}\\
    \caption{Reference-guided synthesis results on the FFHQ dataset.}
    \label{fig:ffhq}
\end{figure*}
}

\newcommand{\figchurch}{
\renewcommand{\h}{33mm}
\renewcommand{\himg}{-0.9mm}
\renewcommand{\hlong}{132mm}
\begin{figure*}[p!]
    \centering
    \includegraphics[width=\h]{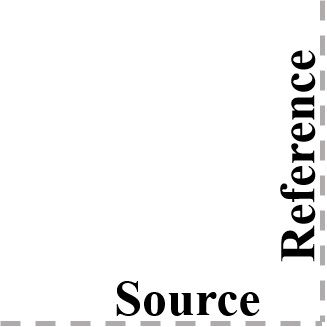}\hspace{\hsrc}
    \includegraphics[width=\hlong]{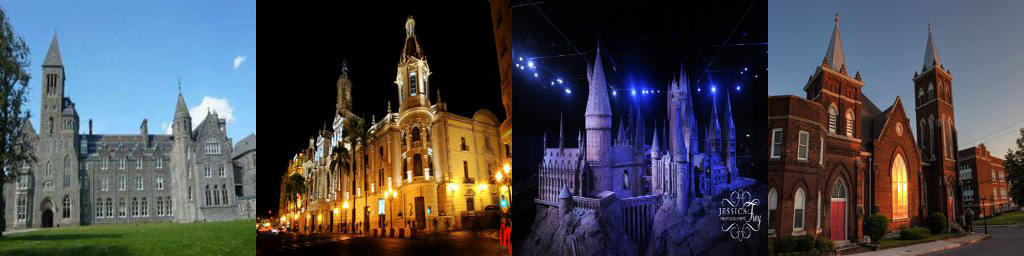}\\
    \includegraphics[width=\h]{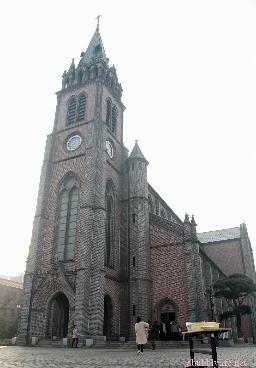}\hspace{\hsrc}
    \includegraphics[width=\hlong]{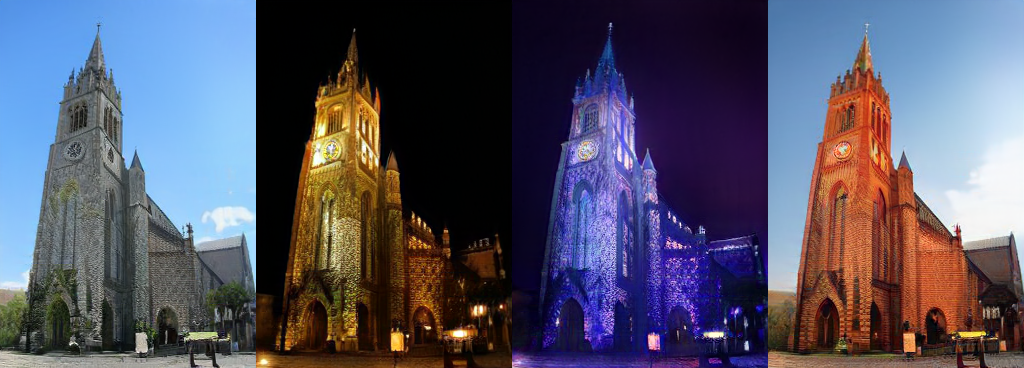}\\
    \includegraphics[width=\h]{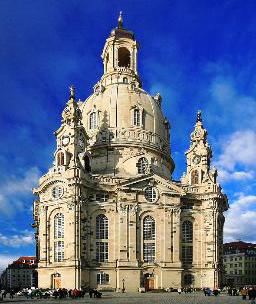}\hspace{\hsrc}
    \includegraphics[width=\hlong]{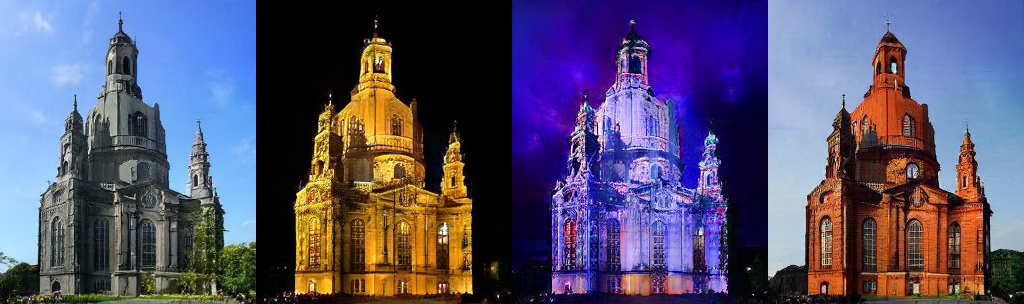}\\
    \includegraphics[width=\h]{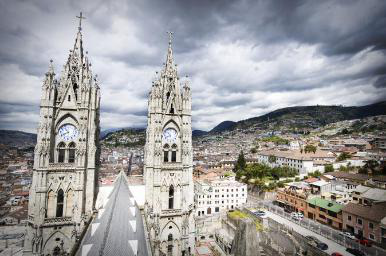}\hspace{\hsrc}
    \includegraphics[width=\hlong]{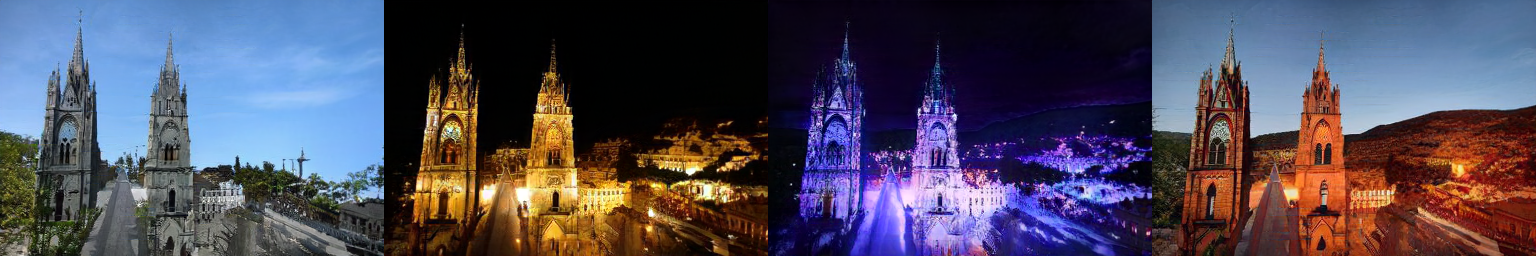}\\
    \caption{Reference-guided synthesis results on the LSUN churches dataset. The model was trained at 256$\times$256 resolution and tested at 256 resolution on the shorter side.}
    \label{fig:church}
\end{figure*}
}

\newcommand{\figflower}{
\renewcommand{\h}{33mm}
\renewcommand{\himg}{-0.9mm}
\renewcommand{\hsrc}{0.1mm}
\begin{figure*}[p!]
    \centering
    \includegraphics[width=\h]{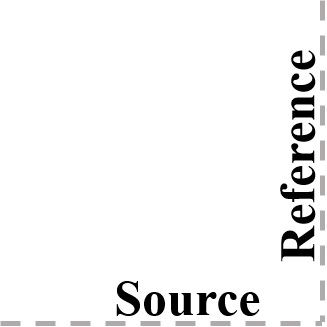}\hspace{\hsrc}
    \includegraphics[width=\h]{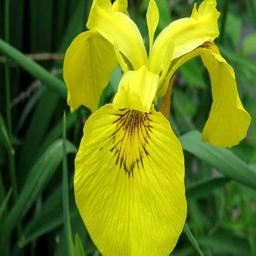}\hspace{\himg}
    \includegraphics[width=\h]{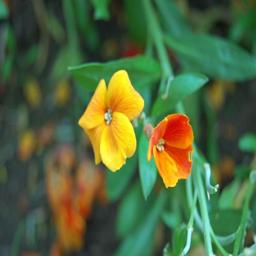}\hspace{\himg}
    \includegraphics[width=\h]{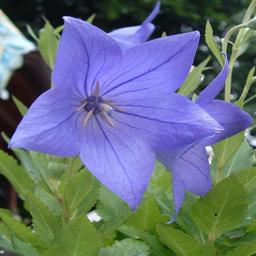}\hspace{\himg}
    \includegraphics[width=\h]{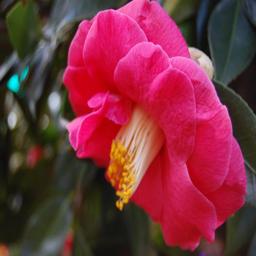}\\\vspace{1mm}
    \includegraphics[width=\h]{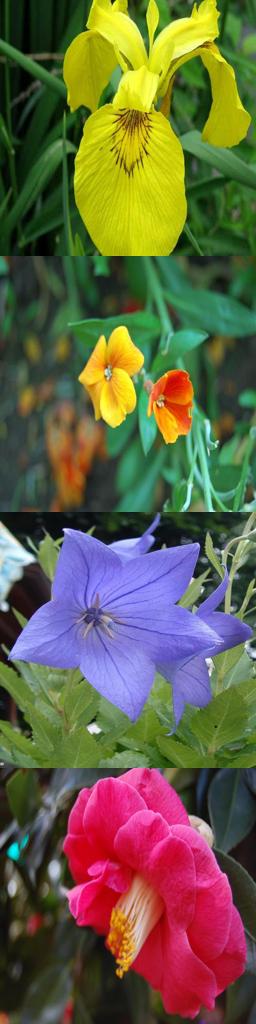}\hspace{\hsrc}
    \includegraphics[width=\h]{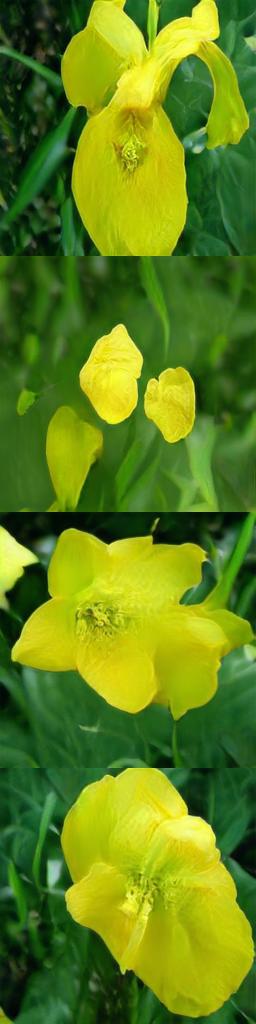}\hspace{\himg}
    \includegraphics[width=\h]{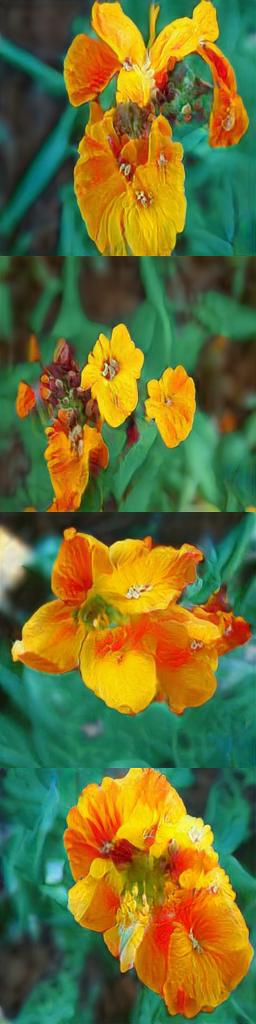}\hspace{\himg}
    \includegraphics[width=\h]{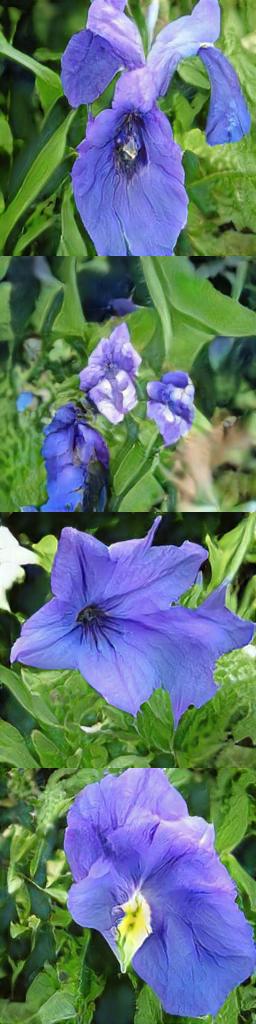}\hspace{\himg}
    \includegraphics[width=\h]{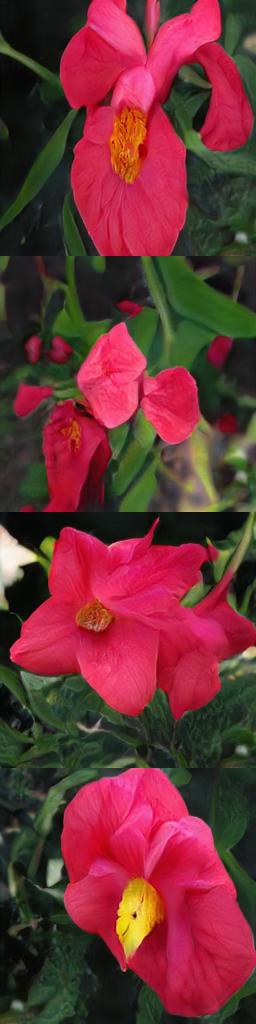}\\
    \caption{Reference-guided synthesis results on the Oxford-102 dataset. The model was trained and tested at 256$\times$256 resolution.}
    \label{fig:flower}
\end{figure*}
}

\newcommand{\figproto}{
\renewcommand{\h}{17mm}
\renewcommand{\hlong}{136mm}
\renewcommand{\himg}{-0.7mm}
\renewcommand{\hsrc}{0.1mm}
\begin{figure*}[p!]
    \centering
    \footnotesize{
        \makebox[\h][c]{\textbf{Source}}\hspace{\hsrc}
        \makebox[\hlong][c]{\textbf{Prototypes}}\hspace{\hsrc}
    }\\
    \includegraphics[width=\h]{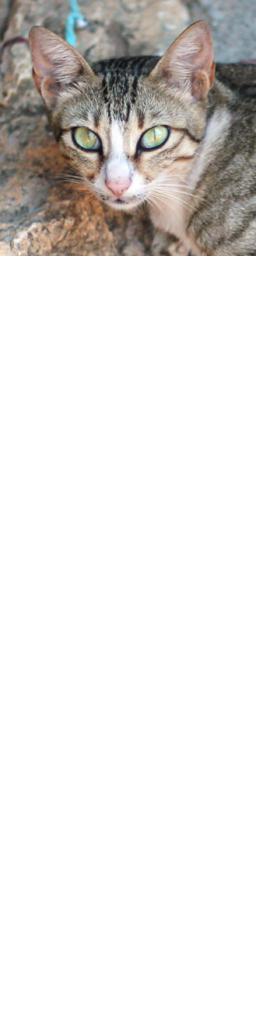}\hspace{\hsrc}
    \includegraphics[width=\hlong]{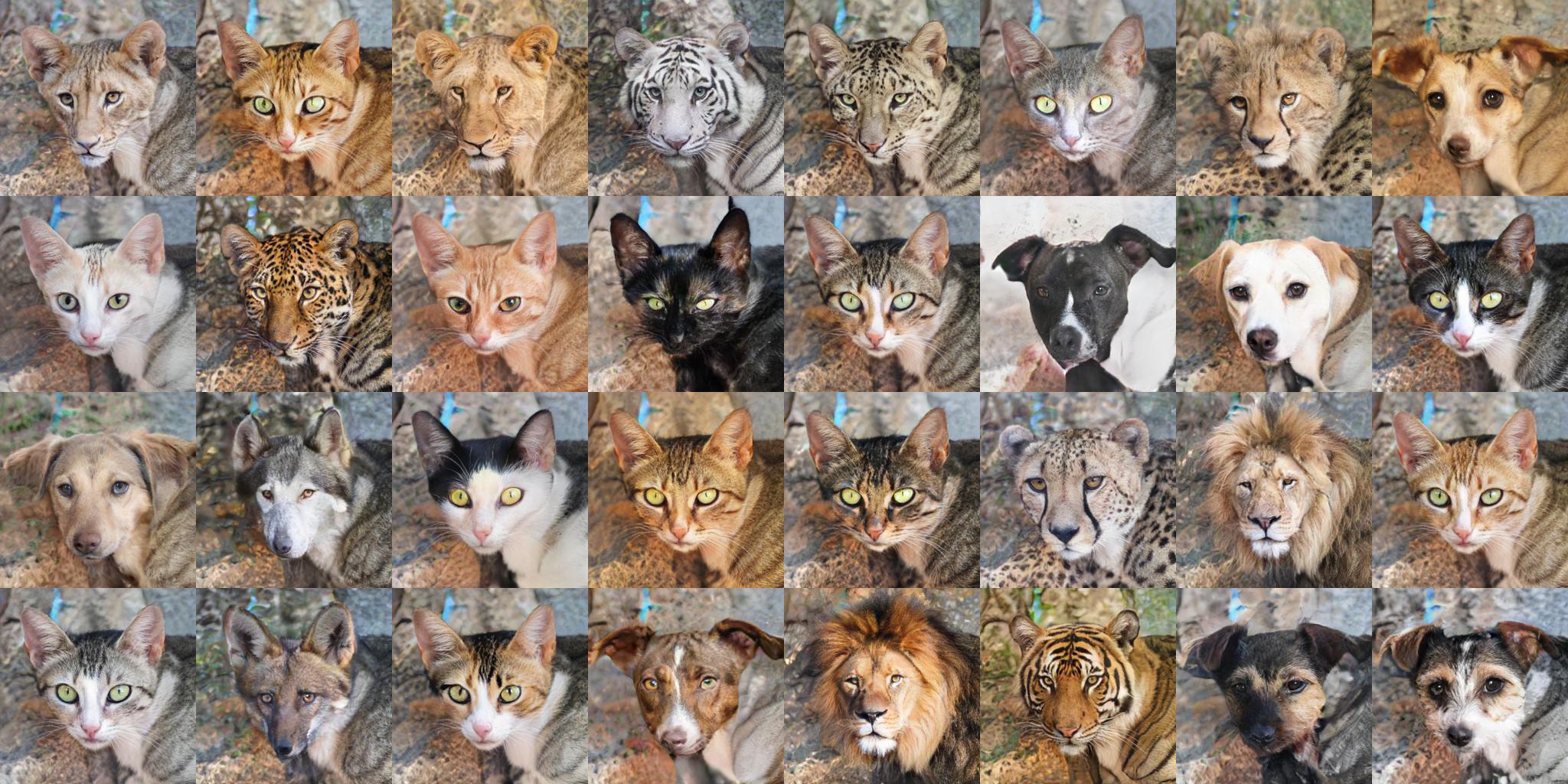}\\
    \footnotesize{
        \makebox[\h][c]{\textbf{Source}}\hspace{\hsrc}
        \makebox[\hlong][c]{\textbf{Prototypes}}\hspace{\hsrc}
    }\\
    \includegraphics[width=\h]{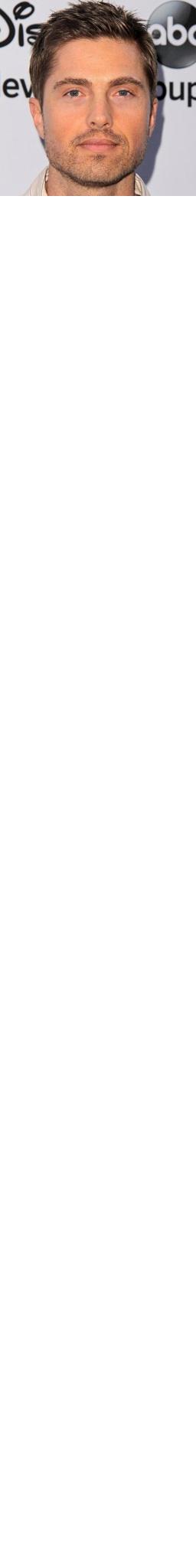}\hspace{\hsrc}
    \includegraphics[width=\hlong]{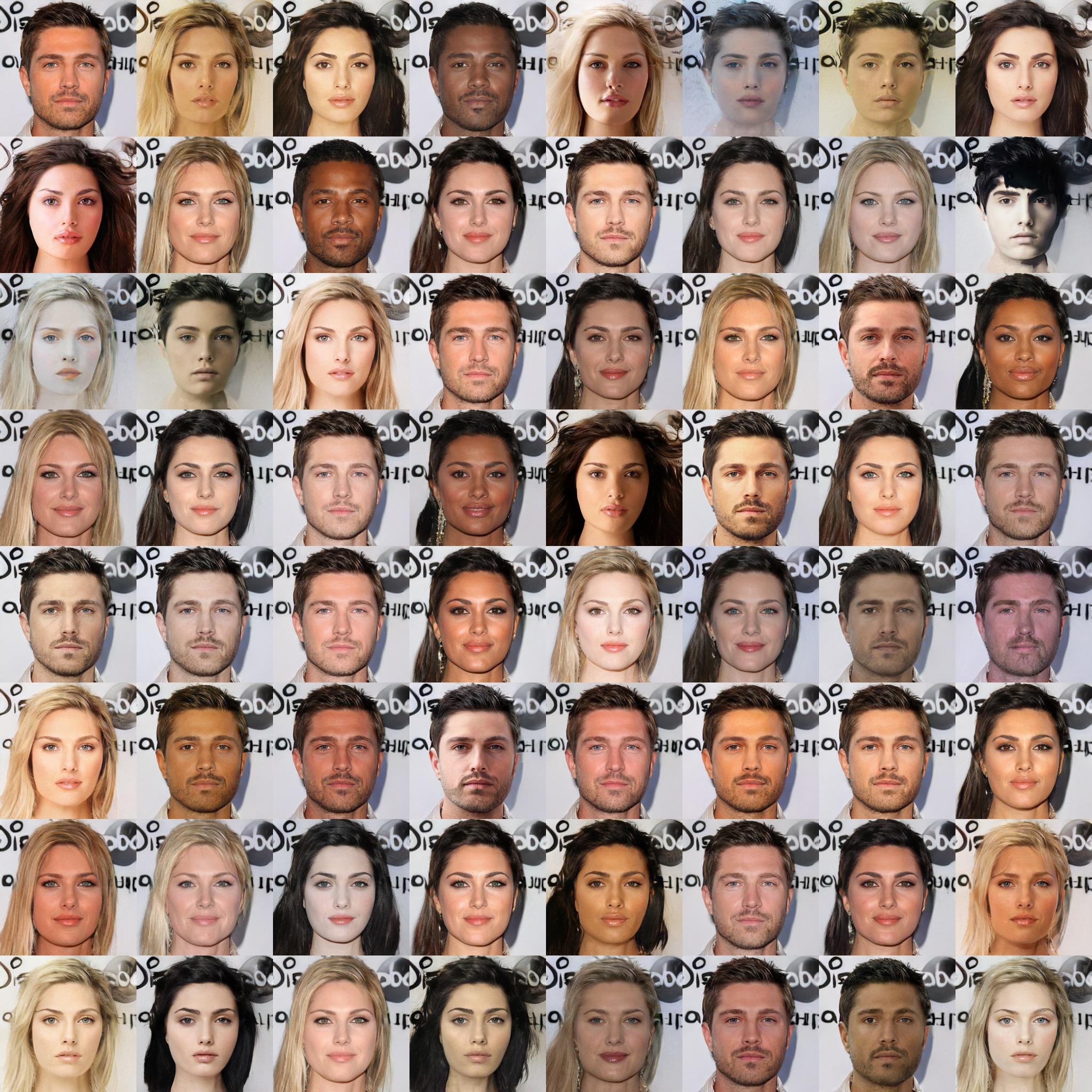}
    \caption{Visualization of all prototypes. (Top) 32 prototypes learned with the AFHQ dataset. (Bottom) 64 prototypes learned with the CelebA-HQ dataset.}
    \label{fig:proto}
\end{figure*}
}

\newcommand{\hlrow}{\rowcolor{blue!7}}

\newcommand{\tabresult}{
\begin{table*}
    \centering
    \begin{tabular}{l c c c c c c c c c}
        \toprule
        & &\multicolumn{4}{c}{{\bf Latent-guided synthesis}} &\multicolumn{4}{c}{{\bf Reference-guided synthesis}}\\\cmidrule(lr){3-6} \cmidrule(lr){7-10}
        & &\multicolumn{2}{c}{\textbf{AFHQ}} &\multicolumn{2}{c}{\textbf{CelebA-HQ}} &\multicolumn{2}{c}{\textbf{AFHQ}} &\multicolumn{2}{c}{\textbf{CelebA-HQ}}\\\cmidrule(lr){3-4} \cmidrule(lr){5-6} \cmidrule(lr){7-8} \cmidrule(lr){9-10}
        \textbf{Method} &Param. (M) &FID$\downarrow$ &KID$\downarrow$ &FID$\downarrow$ &KID$\downarrow$ &mFID$\downarrow$ &mKID$\downarrow$ &mFID$\downarrow$ &mKID$\downarrow$\\\midrule
        \hlrow \ours &\textbf{56.51} &\textbf{\textit{10.0}} &\textbf{2.1} &\textbf{6.8} &\textbf{2.8} &\textbf{10.6} &\textbf{2.1} &\textbf{12.6} &\textbf{4.9}\\
        \hlrow \adain &57.75 &12.5 &2.5 &10.9 &4.9 &14.7 &5.4 &17.6 &8.6\\
        *StarGAN v2 \cite{choi2020stargan} &87.67 &\textbf{9.8} &2.3 &13.9 &8.0 &20.0 &9.8 &28.3 &17.3\\
        *Liu \textit{et al.} \cite{liu2021smoothing} &87.67 &26.0 &7.0 &17.8 &11.0 &51.7 &28.6 &26.7 &16.8\\
        TUNIT \cite{baek2021rethinking} &107.70 &116.1 &99.7 &128.0 &122.0 &223.0 &187.7 &173.7 &193.7\\
        CLUIT \cite{lee2021contrastive} &80.54 &\multicolumn{4}{c}{N/A} &22.6 &10.5  &28.9 &18.1\\\hdline
        SwapAE \cite{park2020swapping} &109.03 &\multicolumn{4}{c}{N/A} &61.2 &28.8 &25.4 &17.8\\
        *StyleMapGAN \cite{kim2021exploiting} &126.23 &32.8 &18.7 &24.3 &15.2 &64.3 &51.3 &28.8 &25.1\\\bottomrule
    \end{tabular}
    \vspace{-1mm}
    \caption{Quantitative comparison on image synthesis. We report FID and $\text{KID}\times10^3$. An asterisk (*) denotes that we used the pre-trained networks provided by authors. \textbf{Bold} indicates the best result and \textbf{\textit{bold}+\textit{italicize}} indicates the best result among the \textit{unsupervised} methods.}
    \vspace{-2mm}
    \label{tab:results}
\end{table*}
}

\newcommand{\tabuserstudy}{
\begin{table}
    \centering
    \setlength{\tabcolsep}{5pt}
    \begin{tabular}{l c c c c c c}\toprule
        &\multicolumn{3}{c}{\textbf{AFHQ} (\%)} &\multicolumn{3}{c}{\textbf{CelebA-HQ} (\%)}\\\cmidrule(lr){2-4} \cmidrule(lr){5-7}
        Method &Q1 &Q2 &Q3 &Q1 &Q2 &Q3\\\midrule
        \hlrow \ours &\textbf{24.5} &\textbf{22.4} &\textbf{25.0} &\textbf{\textit{25.0}} &\textbf{\textit{19.3}} &\textbf{\textit{23.2}}\\
        CLUIT &18.7 &19.7 &18.1 &14.2 &15.0 &15.5 \\
        TUNIT &21.9 &18.0 &17.9 &11.8 &10.7 &9.2 \\\hdline
        $\star$Liu \textit{et al.} &19.3 &19.2 &20.1 &\textbf{26.7} &\textbf{28.2} &\textbf{26.3}\\
        $\star$StarGAN v2 &15.6 &20.6 &18.8 &22.4 &26.8 &25.7\\\bottomrule
    \end{tabular}
    \vspace{-1mm}
    \caption{User study. Q1: content and style. Q2: realism. Q3: preference. A star ($\star$) denotes models trained with extra information.}
    \label{tab:userstudy}
\end{table}
}

\newcommand{\tabproj}{
\begin{table}
    \centering
    \setlength{\tabcolsep}{3.5pt}
    \begin{tabular}{l c c c c c c}\toprule
        & &\multicolumn{2}{c}{\textbf{AFHQ}} &\multicolumn{2}{c}{\textbf{CelebA-HQ}}\\\cmidrule(lr){3-4} \cmidrule(lr){5-6}
        \textbf{Method} &\multirow{-2}{*}{\shortstack[c]{Runtime \\ (sec)}} &MSE &LPIPS &MSE &LPIPS\\\midrule
        \hlrow \ours &\textbf{0.029} &0.012 &\textbf{0.269} &0.007 &\textbf{0.202}\\
        SwapAE &0.037 &\textbf{0.009} &0.303 &\textbf{0.005} &0.241\\
        StyleMapGAN &0.092 &0.039 &0.316 &0.026 &0.255\\\bottomrule
    \end{tabular}
    \vspace{-1mm}
    \caption{Quantitave comparision for real image projection. We used a single NVIDIA Xp GPU to measure the runtime.}
    \vspace{-3mm}
    \label{tab:projection}
\end{table}
}

\newcommand{\tabcombine}{
\begin{table}
    \centering
    \setlength{\tabcolsep}{7pt}
    \begin{tabular}{l c c c}\toprule
        Method &Param. (M) &k-NN$\uparrow$ &mFID$\downarrow$\\\midrule
        \hlrow \adain &\textbf{57.75} &\textbf{99.1} &\textbf{14.7}\\
        \texttt{separated} &76.77 &80.6 &159.7\\\bottomrule
    \end{tabular}
    \vspace{-1mm}
    \caption{Quantitative comparison using the AFHQ dataset.}
    \vspace{-2mm}
    \label{tab:combine}
\end{table}
}

\newcommand{\tabmemory}{
\begin{table}
    \vspace{-2mm}
    \centering
    \begin{tabular}{l c c c c}\toprule
        &\multicolumn{3}{c}{Parameters (M)} &\multirow{2}{*}{sec/iter}\\\cmidrule(lr){2-4}
        Method &$G$ &$D$ &$E$\\\midrule
        \hlrow \ours &36.8 &\multicolumn{2}{c}{\textbf{19.7}} &0.383\\  
        \hlrow \adain &38.1 &\multicolumn{2}{c}{\textbf{19.7}} &\textbf{0.351}\\  
        StarGAN v2 \cite{choi2020stargan} &43.5 &20.9 &20.9 &0.678\\                
        TUNIT \cite{baek2021rethinking} &\textbf{27.4} &71.0 &9.3 &0.667\\   
        CLUIT \cite{lee2021contrastive} &34.4 &25.2 &20.9 &1.016\\\hdline    
        SwapAE \cite{park2020contrastive} &25.1 &53.4 &30.6 &0.692\\         
        StyleMapGAN \cite{kim2021exploiting} &79.7 &28.9 &17.6 &1.475\\\bottomrule  
    \end{tabular}
    \vspace{-1mm}
    \caption{Efficiency of proposed method. We measured the training speed (s/iter) with minibatch size 2 on a single TITAN Xp GPU.}
    \vspace{-2mm}
    \label{tab:memory}
\end{table}
}

\newcommand{\tabunlabel}{
\begin{table}
    \centering
    \setlength{\tabcolsep}{9pt}
    \begin{tabular}{l c c}\toprule
        &\multicolumn{2}{c}{FID}\\ \cmidrule{2-3}
        Method &\textbf{Churches} &\textbf{FFHQ $256^2$}\\ \midrule
        \hlrow \textbf{Ours} (latent) &\textbf{9.0} &5.2\\
        \hlrow \textbf{Ours} (reference) &12.2 &\textbf{5.1}\\
        *SwapAE \cite{park2020swapping} &49.6 &-\\\hdline
        StyleGAN2 \cite{karras2020analyzing} &4.1 &3.7\\\bottomrule
    \end{tabular}
    \vspace{-1mm}
    \caption{Quantitative comparison using the unlabeled datasets. An asterick (*) indicates that we used the pre-tranined networks provided by the authors. Note that we calculated StyleGAN2 results using randomly sampled images, not manipulated images (\ie style mixing).}
    \label{tab:unlabel}
\end{table}
}

\newcommand{\tablerp}{
\begin{table}
    \centering
    \setlength{\tabcolsep}{11pt}
    \begin{tabular}{l c c}\toprule
        &\multicolumn{2}{c}{$\textrm{FID}_{lerp}$}\\\cmidrule{2-3}
        Method &\textbf{AFHQ} &\textbf{CelebA-HQ}\\\midrule
        \hlrow \ours &\textbf{11.2} &\textbf{25.4}\\
        \hlrow \adain &\textbf{\textit{14.0}} &\textbf{\textit{31.0}}\\
        Liu \etal \cite{liu2021smoothing} &30.0 &35.8\\
        StarGAN v2 \cite{choi2020stargan} &32.2 &76.8\\\bottomrule
    \end{tabular}
    \vspace{-1mm}
    \caption{Quantitative comparison of the style interpolation.}
    \label{tab:lerp}
\end{table}
}

\newcommand{\tababalk}{
\begin{table}
    \centering
    \begin{tabular}{r c c c c c c} \toprule
        $K$ &32 &64 &128 &256 &512 &1024\\\midrule
        k-NN$\uparrow$ &99.1 &99.1 &98.8 &98.5 &96.3 &95.6\\
        mFID$\downarrow$ &14.7 &15.8 &28.4 &26.6 &34.6 &42.1\\\bottomrule
    \end{tabular}
    \vspace{-1mm}
    \caption{Effect of the number of prototypes. Note that mFID of supervised method (StarGAN v2) \cite{choi2020stargan} is 24.1.}
    \label{tab:abalk}
\end{table}
}

\definecolor{codegreen}{rgb}{0,0.6,0}
\definecolor{comment}{rgb}{0,0.3,0.7}
\definecolor{codegray}{rgb}{0.5,0.5,0.5}
\definecolor{codepurple}{rgb}{0.58,0,0.82}
\definecolor{backcolour}{rgb}{0.95,0.95,0.92}
\lstdefinestyle{mystyle}{
    commentstyle=\color{comment},
    keywordstyle=\color{magenta},
    numberstyle=\tiny\color{codegray},
    stringstyle=\color{codepurple},
    basicstyle=\ttfamily\footnotesize,
    breakatwhitespace=false,         
    breaklines=true,                 
    captionpos=b,                    
    keepspaces=true,                 
    numbers=left,                    
    numbersep=2pt,                  
    showspaces=false,                
    showstringspaces=false,
    showtabs=false,                  
    tabsize=4
}
\usepackage[pagebackref,breaklinks,colorlinks]{hyperref}

\usepackage[capitalize]{cleveref}
\crefname{section}{Sec.}{Secs.}
\Crefname{section}{Section}{Sections}
\Crefname{table}{Table}{Tables}
\crefname{table}{Tab.}{Tabs.}

\def\cvprPaperID{10144} 
\def\confName{CVPR}
\def\confYear{2022}

\begin{document}

\title{A Style-aware Discriminator for Controllable Image Translation}

\author{Kunhee Kim \quad Sanghun Park \quad Eunyeong Jeon \quad Taehun Kim \quad Daijin Kim\\
Pohang University of Science and Technology (POSTECH)\\
{\tt\small \{kunkim,sanghunpark,eyjeon,taehoon1018,dkim\}@postech.ac.kr}
}
\maketitle

\begin{abstract}
   Current image-to-image translations do not control the output domain beyond the classes used during training, nor do they interpolate between different domains well, leading to implausible results. This limitation largely arises because labels do not consider the semantic distance. To mitigate such problems, we propose a style-aware discriminator that acts as a critic as well as a style encoder to provide conditions. The style-aware discriminator learns a controllable style space using prototype-based self-supervised learning and simultaneously guides the generator. Experiments on multiple datasets verify that the proposed model outperforms current state-of-the-art image-to-image translation methods. In contrast with current methods, the proposed approach supports various applications, including style interpolation, content transplantation, and local image translation. The code is available at \href{https://github.com/kunheek/style-aware-discriminator}{github.com/kunheek/style-aware-discriminator}.
\end{abstract}

\section{Introduction}
\label{sec:intro}
Image-to-image (I2I) translation aims to manipulate the style of an existing image, where style refers to generic attributes that can be applied to any image in a dataset (\eg, texture or domain). Content generally refers to the remaining information, such as the pose and structure. This task has shown significant progress with generative adversarial network (GAN) \cite{goodfellow2014gan} developments. Recent studies have expanded the functionality to multi-modal and multi-domains using domain-specific discriminators and latent injection \cite{huang2017arbitrary}, enabling the direct manipulation of existing images using domain labels or reference images \cite{choi2018stargan,choi2020stargan,liu2019few,saito2020coco,liu2021smoothing}.

However, despite promising functionality advances, there remains considerable room for development in terms of controllability. For example, users can only control the classes used for training. Although a reference image can be used to control output but this can often lead to erroneous results, particularly for misrecognition within the same class; and another common problem is inability to fine-tune the output. Since the label space does not consider the semantic distance between classes, the learned style space cannot reflect these semantic distances, which leads to unrealistic images when controlling the results by manipulating the style code \cite{liu2021smoothing}.

This study investigates I2I translation controllability, \ie, to be able to edit the result as desired using the style code, without being limited to the previously defined label space. The proposed model learns the style space using prototype-based self-supervised learning \cite{caron2020unsupervised} with carefully chosen augmentations. Although the current domain-specific discriminators are not designed for an external continuous space, this is possible if the discriminator knows the style internally. Therefore, we propose a \textit{Style-aware Discriminator}, combining a style encoder and a discriminator into a single module. Thus, the proposed model is somewhat lighter by reducing one module and achieves better performance because of to the better representation space of the discriminator. We used the style code sampled from prototypes during training to improve the controllability; and feature-level and pixel-level reconstructions to improve the consistency. Thus, the proposed model goes beyond image translation to support various applications, including style interpolation and content transplantation. Finally, we propose feedforward local image translation by exploiting spatial properties of the GAN feature space.

We evaluated the model on several challenging datasets: Animal Faces HQ (AFHQ) \cite{choi2020stargan}, CelebA-HQ \cite{karras2018progressive}, LSUN churches \cite{yu2015lsun}, Oxford-102 \cite{Nilsback08}, and FlickrFaces-HQ (FFHQ) \cite{karras2019style}. Extensive experiments confirm that the proposed method outperforms current state-of-the-art models in terms of both performance and efficiency without semantic annotations. The proposed model can also project an image into the latent space faster than baselines while achieving comparable reconstruction results.

The contributions from this study are summarized as follows: (i) We propose an integrated module for style encoding and adversarial losses for I2I translation, as well as a data augmentation strategy for the style space. The proposed method reduces the parameter count significantly and does not require semantic annotations. (ii) We achieve state-of-the-art results in \textit{truly} unsupervised I2I translation in terms of the Fr\'{e}chet Inception Distance (FID) \cite{heusel2017fid}. The proposed method shows similar or better performance compared with supervised methods. (iii) We extend image translation functionality to various applications, including style interpolation, content transplantation, and local image translation.

\section{Related work}
\label{sec:related}
\noindent\textbf{Multi-domain I2I translation}\quad StarGAN \cite{choi2018stargan} enabled many-to-many translation using a given attribute label, but this and similar approaches have the disadvantage of being deterministic for a given input and domain. Subsequent studies suggested using reference images rather than labels \cite{liu2019few,saito2020coco}, enabling translation based on an image from unseen classes in the same domain. StarGAN v2 \cite{choi2020stargan} introduced a noise-to-latent mapping network to synthesize diverse results for the same domain, but since all of these methods depend on labels defined for classification, representations for image manipulation cannot be learned. Therefore, we developed a new multi-domain I2I approach from two perspectives. The proposed method learns a style-specific representation suitable for image manipulation without relying on labels; and then provides more user-controllability while supporting various applications.

To overcome the problem of label dependency, Bahng \etal \cite{bahng2020exploring} clustered the feature space for pre-trained networks to create pseudo-labels and corresponding latent code. Similarly, TUNIT trained a guiding network using contrastive learning and clustering directly in target data \cite{baek2021rethinking}. These methods obtain pseudo-labels that can be substituted for class labels; however, the proposed approach models a continuous style space rather than discrete pseudo-labels. Thus the proposed model is significantly more efficient than previous approaches. CLUIT \cite{lee2021contrastive} recently proposed using contrastive learning through a discriminator, but used contrastive learning to replace the multi-task discriminator. Therefore, the style encoder exists independently, in contrast with the proposed model. Furthermore, CLUIT requires additional process (\eg, clustering), to obtain a style code without a reference image.

\noindent\textbf{Learning-based image editing}\quad Recently, Karras \etal discovered that GANs naturally learn to disentangle pre-defined latent space \cite{karras2019style,karras2020analyzing}. Several subsequent sutides proposed image editing methods using StyleGANs \cite{abdal2019image2stylegan,abdal2020image2stylegan++,zhu2020domain}. However, these methods suffered from the long time it takes to find a corresponding latent. Recently, StyleMapGAN \cite{kim2021exploiting} and Swapping Autoencoder (SwapAE) \cite{park2020swapping} proposed directly encoding an image into the latent, enabling real-time and various image editing applications. Our study is different in that content and style can be manipulated separately because of disentangled latent space. SwapAE has a separate latent space called texture and structure, similar to the proposed method, but is challenging to operate without a reference image. In addition, its texture-focused representation does not work well for tasks that require dramatic changes, such as the interspecies variation of animal faces (Fig.~\ref{fig:reference}). On the other hand, since the proposed method learns the style space and the prototype, manipulating an image without a reference image is possible. Furthermore, because our method is designed for I2I translation, more challenging manipulations are possible.

\noindent\textbf{Discriminator and self-supervised learning}\quad GANs \cite{goodfellow2014gan} have always suggested that discriminators could be feature extractors, and many previous studies have demonstrated that GANs benefit from representation learning through a discriminator \cite{chen2019self,li2021jigsawgan,liu2020towards,jeon2021fa,jeong2020training}. We also utilize self-supervised learning via the discriminator, but differ from previous approaches in that the primary purpose of the self-supervised learning is to function as an encoder, not just to improve the quality. Hence, our discriminator continues to work as an encoder after training; as opposed to most current GANs, which abandon discriminators after training.

\section{Methods}
\label{sec:methods}
\figframework
Our aim was to build a flexible and manipulative style space. In particular, we considered the following objectives.
(i) Visual similarity should be considered. For example, visually similar pairs, such as a wolf and a dog, should be placed in similar places in the style space.
(ii) There should be a representative value, such as a discrete label, providing a good starting point when the user wants to fine-tune the result.
\subsection{Framework}
The framework overview is shown in Fig.~\ref{fig:framework}, which shows the sampling strategies for the style code and the training procedures of the entire model.

\noindent\textbf{Style-aware discriminator}\quad Given an image $\textbf{x} \in \mathcal{X}$, discriminator $D$ returns a vector as output. The discrimination head $h_D$ determines whether $\textbf{x}$ is a real or fake image, and the style head outputs the latent code $\textbf{z}_s = h_s(D(\textbf{x}))$. We formulate the traditional discriminator $f_D(\textbf{x})$ and the style encoder $f_s(\textbf{x})$ as $h_D(D(\textbf{x}))$ and $h_s(D(\textbf{x}))$, respectively.

\noindent\textbf{Prototypes}\quad We represent the style space using a set of L2-normalized vectors $\textbf{C} \in \mathbb{R}^{K\times D}$ rather than predefined labels or pseudo-labels, where $K$ and $D$ denote the number of prototypes and style code dimension, respectively. We denote $\textbf{c}_k$ as an element of $\textbf{C}$.

\noindent\textbf{Generator}\quad The generator comprises an encoder and a decoder, similar to typical current I2I translation generators \cite{choi2020stargan}. The encoder $G_{enc}(\textbf{x})$ extracts style-agnostic content code $\textbf{z}_c \in \mathbb{R}^{D\times W \times H}$ from input $\textbf{x}$, and decoder $G_{dec}(\textbf{z}_c,\textbf{z}_s)$ synthesizes a new image reflecting content code and style code. Similar to Karras \etal \cite{Karras2020training}, we use a 2-layer multi-layer perceptron (MLP) to transform the normalized style code $\textbf{z}_s$ into valid features. The generator uses weight modulation \cite{karras2020analyzing} or AdaIN \cite{huang2017arbitrary} for latent injection.

\subsection{Modeling the style space}
Intuitively, an ideal style encoder would output the same code even though the input image was geometrically transformed. This idea is the fundamental concept underlying contrastive learning \cite{oord2018representation,chen2020simple,he2020momentum,li2020prototypical,caron2020unsupervised}, which has been actively studied in recent years. We adopted self-supervised learning in our framework to learn the style space.

\noindent\textbf{Data augmentation}\quad The goal of existing contrastive learning is to classify object instances other than the style in images. Chen \etal \cite{chen2020simple} proposed a specific augmentation pipeline (\eg, random crop, color distortion) which has become the preferred approach. However, distorting the color does not serve our purpose since style is deeply related to color. Hence we use geometric transforms (\eg, scale, rotation) to learn content invariant representation, and cutout \cite{devries2017improved} to learn styles such as gender and facial expressions for human faces. We also use random crop and resize following the work of \cite{chen2020simple}.

\noindent\textbf{SwAV}\quad We used the SwAV framework \cite{caron2020unsupervised}, online clustering based self-supervised learning, because it aligns with our goals in terms of updating prototypes and achieving better performance for small batch sizes. The basic concept is that encoded representations from both views (\ie, augmented images) for the same image predict each other's assignments $\textbf{q}$. The objective for learning style space is expressed as:
\begin{equation}
    \mathcal{L}_{swap} = l(\textbf{q}^{(2)}, \textbf{z}_s^{(1)}) + l(\textbf{q}^{(1)}, \textbf{z}_s^{(2)}),
\end{equation}
where $l(\textbf{q}, \textbf{z}_s) = -\sum_{k}^{K} \textbf{q}_k (\exp(\frac{\textbf{z}_s\cdot \textbf{c}_k}{\tau})/\sum_{k'}^K\exp(\frac{\textbf{z}_s\cdot \textbf{c}_{k'}}{\tau}))$, $\tau$ is a temperature parameter, and $\textbf{q}$ is a code computed using the Sinkhorn algorithm \cite{cuturi2013sinkhorn,caron2020unsupervised}. Note that swapped prediction loss can be replaced by other self-supervised learning objectives, such as InfoNCE \cite{oord2018representation}, by sacrificing the advantages of the prototype.

\subsection{Learning to synthesize}
During training, we sample a target style code $\tilde{\textbf{z}_s}$ from the prototype or dataset $\mathcal{X}$. When sampling from the prototype, we use perturbed prototypes or samples that are linearly interpolated between two prototypes (see Appendix~\ref{appx:a3} for more details). Then, we apply a stop-gradient to prevent the style space from being affected by other objectives.

As shown in Fig.~\ref{fig:framework} (c), the generator $G$ synthesizes a fake image $G(\textbf{x}, \tilde{\textbf{z}_s})$. To enforce synthesized image be realistic, we adopted a non-saturating adversarial loss \cite{goodfellow2014gan}:
\begin{equation}
    \mathcal{L}_{adv} = \mathbb{E}_{\textbf{x}} \left[ \log(f_D(\textbf{x})) \right] +
    \mathbb{E}_{\textbf{x},\tilde{\textbf{z}_s}} \left[ \log(1 - f_D(G(\textbf{x}, \tilde{\textbf{z}_s}))) \right].
\end{equation}
We also employed R1 regularization \cite{mescheder2018r1reg} following previous works \cite{choi2020stargan,baek2021rethinking,lee2021contrastive,park2020swapping,kim2021exploiting}.

We adopted a \textit{style reconstruction loss} to ensure the generator $G$ utilize the style code:
\begin{equation}
    \mathcal{L}_{style} = \mathbb{E}_{\textbf{x},\tilde{\textbf{z}_s}}[||\tilde{\textbf{z}_s} - f_{s}(G(\textbf{x}, \tilde{\textbf{z}_s})) ||_2^2],
\end{equation}
Previous multi-domain and multi-modal I2I translation methods \cite{choi2020stargan,huang2018munit,baek2021rethinking} introduced similar objectives, the difference between the current and previous approaches is that we do not update a style encoder using this objective.

\subsection{Disentanglement of style and content}
An ideal image manipulation network should be able to separate an image into two mutually exclusive representations and synthesize them back into the original image without information loss \cite{huang2018munit}. Thus, the framework must satisfy the following:
\begin{equation}
    \phi(\textbf{x}, G(f_c(\textbf{x}), f_s(\textbf{x}))) = 0,
\end{equation}
where $\phi(\cdot)$ is a distance measure in pixel space; and $f_c(\textbf{x})$, $f_s(\textbf{x})$ are encoding functions for content and style, respectively.
To achieve this, we employ a \textit{reconstruction loss}:
\begin{equation}
    \mathcal{L}_{recon} = \mathbb{E}_{\textbf{x}} \left[ \phi(\textbf{x}, G(\textbf{x}, \texttt{sg}(f_s(\textbf{x})))) \right],
\end{equation}
where \texttt{sg} denotes a stop-gradient operation. This objective encourages $G_{enc}$ to encode mutually exclusive features with the style code since $f_s(\textbf{x})$ is not updated. Although any distance measure in pixel space can be used, we used learned perceptual image patch similarity (LPIPS) \cite{zhang2018unreasonable} since we empirically found this works better than Euclidean or Manhattan distance.

In order to learn content space through the reconstruction loss above, it is necessary to condition that the generator should not ignore input latents code. For example, the generator may ignore the content code and perform reconstruction with only style code. To prevent this, we enforce the generator to preserve input content code using a \textit{content reconstruction loss}:
\begin{equation}
    \mathcal{L}_{content} = \mathbb{E}_{\textbf{x},\tilde{\textbf{z}_s}}\left[\frac{1}{WH}\sum_{i,j}^{W,H} ||\textbf{z}_{c,i,j} - \tilde{\textbf{z}}_{c,i,j}||_2^2 \right],
\end{equation}
where $\textbf{z}_c$, $\tilde{\textbf{z}_c}$ are $G_{enc}(\textbf{x})$, $G_{enc}(G(\textbf{z}_c,\tilde{\textbf{z}_s}))$, respectively.
This objective enforces patch-level similarity between inputs and outputs, similar to PatchNCE \cite{park2020contrastive}. However, our proposed objective is simpler since we only compare the last layer features, and our objective does not contrast features between patches.

In practice, we found that there was no need to apply this loss every step, and hence we apply the objective every 16th step. We assume that this is because similar results can be obtained through a \textit{reconstruction loss}.

\noindent\textbf{Overall objectives}\quad Our final objective function for the discriminator is $\mathcal{L}_{StyleD} = \mathcal{L}_{adv} + \lambda_{swap} \mathcal{L}_{swap}$,
and for the generator is $\mathcal{L}_{G} = \mathcal{L}_{adv} + \lambda_{sty} \mathcal{L}_{style} + \lambda_{rec} \mathcal{L}_{recon}$,
where $\lambda_{sty}, \lambda_{rec}$ are hyperparameters for each term, and we use for all $\lambda = 1.0$ except $\lambda_{rec} = 0.3$ for AdaIN-based models. We set $K$ as 32 and 64 for AFHQ and CelebA-HQ, respectively. Please refer to Appendix~\ref{appx:a} for more details.

\subsection{Local image translation}
One advantage of factored representations is having a higher degree of freedom when editing an image. The content of an image can easily be copied or moved by editing in the content space \cite{park2020swapping}. To progress further, we propose a simple method of patch-level image translation. Kim \etal \cite{kim2021exploiting} proposed mixing spatial information in the latent space to enable local editing. Similarly, we mix spatial information in the feature space.
\begin{equation}
    \textbf{f}_{o} 
        = \textbf{m} \otimes \textrm{\texttt{mod}}(\textbf{f}_i, \textbf{z}_s^{(i)})
        + (1 - \textbf{m}) \otimes \textrm{\texttt{mod}}(\textbf{f}_i, \textbf{z}_s^{(j)}),
\end{equation}
where $\textbf{f}$ and $\textbf{m}$ are feature map and mask, and \texttt{mod} is modulated convolution \cite{karras2020analyzing} or AdaIN. For patch-level image translation, we simply replace the entire modulated convolution layer \cite{karras2020analyzing} with above. To ensure content is maintained even when several styles are mixed, we mixed two styles with a random mask when calculating a content preserving loss.
\section{Experiments}
\label{sec:exp}
\figlatent
\subsection{Experimental setup}
We not only employed a StyleGAN2-based generator but also considered models using AdaIN to enable a fair comparison with I2I translation models that use AdaIN.

\noindent\textbf{Datasets}\quad We trained the proposed and various comparator models on AFHQ, AFHQ v2 \cite{choi2020stargan}, CelebA-HQ \cite{karras2018progressive}, FFHQ \cite{karras2019style}, Oxford-102 \cite{Nilsback08}, and LSUN churches \cite{yu2015lsun}. Since high resolution models requires considerable training time, the proposed and comparison models were trained and evaluated at 256$\times$256 resolution. For AFHQ and CelebA-HQ, we used the splits provided by Choi \etal \cite{choi2020stargan}.

\noindent\textbf{Baselines}\quad Our primary goal is to synthesize an image with a reference image or a latent sampled from a learned space (\ie, I2I translation). We compared the proposed approach with recent supervised \cite{choi2020stargan,liu2021smoothing} and unsupervised \cite{baek2021rethinking,lee2021contrastive} methods. In contrast with most I2I translation methods, the proposed approach has further applications such as image editing. To compare real-time image editing capability, we compared our approach with Swapping Autoencoder (SwapAE) \cite{park2020swapping} and StyleMapGAN \cite{kim2021exploiting}.

We used pre-trained networks provided by the authors whenever possible. Otherwise, we trained the models from scratch using the official implementation, except for CLUIT, where we employed our implementation because the authors have not yet published their code. We showed 1.6 and 5 M images to the AdaIN- and StyleGAN2-based models, respectively. For StyleMapGAN, we used pre-trained networks trained for 5 M images.

\subsection{Main results}
We quantitatively and qualitatively evaluated the proposed approach and the baselines on two datasets: AFHQ and CelebA-HQ.

\noindent\textbf{Latent-guided image synthesis}\quad We report Fr\'{e}chet Inception Distance (FID) \cite{heusel2017fid} and Kernel Inception Distance (KID) \cite{binkowski2018demystifying} to evaluate the latent-guided image synthesis quality, calculating FID and KID between 50,000 synthesized images and training samples. Parmer \etal \cite{parmar2021cleanfid} recently demonstrated that values of these metrics depend on the resizing method; therefore, we calculated FID and KID for all methods using Pillow-bicubic \cite{clark2015pillow}.

To synthesize images, we used a style code sampled using the strategy used in the training. To evaluate supervised methods \cite{choi2020stargan,liu2021smoothing}, we created a style code using randomly sampled domain and noise. We performed style mixing with randomly sampled latent with StyleMapGAN \cite{kim2021exploiting}. In Table~\ref{tab:results}, the proposed model showed better results than the existing unsupervised methods and comparable results to the supervised methods. Although the result of the proposed approach is slightly worse than StarGAN v2 in AFHQ, our approach allows users to choose one of several prototypes, whereas StarGAN v2 only allows users to choose from three classes. In Fig.~\ref{fig:latent}, we show the prototype-guided synthesis results of our methods trained on unlabeled datasets. Note that we directly used prototypes obtained during the training without additional processing.

\figref
\tabresult
\noindent\textbf{Reference-guided image synthesis}\quad Although FID/KID protocol can estimate the manipulated image quality, it provides good performance scores even if the generator ignores the given latent (\eg, reconstruction). Therefore, we evaluated reference-guided image synthesis to evaluate whether the generator reflects the latent corresponding to each domain. Following \cite{choi2020stargan}, we synthesize images using a source-reference pair from each task (\eg, cat$\rightarrow$dog, male$\rightarrow$female) and calculate FID and KID with a training set of a target domain. We report average values of all tasks (mFID and mKID).

As shown in the first two rows of Fig.~\ref{fig:reference}, supervised approaches \cite{choi2020stargan,liu2021smoothing} often misrecognized the style of reference images within the same classes. However, the proposed method successfully captures the styles of reference images. Furthermore, while other methods failed to preserve the details of the source image, the proposed method was the only method that preserved details such as pose and background.

\tabuserstudy
\noindent\textbf{User study}\quad To investigate the human preferences, we conducted a survey using the Amazon MTurk platform. We randomly generated 100 source-reference pairs per dataset and asked the respondents to answer three questions: (Q1) Which one best reflects the style of the reference while preserving the content of the source? (Q2) Which one is the most realistic? (Q3) Which one would you use for manipulating an image? Each set was answered by 10 respondents. As shown in Table~\ref{tab:userstudy}, the respondents obviously preferred our method in the AFHQ. In the CelebA-HQ, our model was not preferred over the supervised models (which use attribute labels and a pre-trained face alignment network); nevertheless, our model was still the most preferred among the unsupervised methods.

See Appendix~\ref{appx:b} for additional results including experiments on AFHQ v2 and Oxford-102.

\figcontrol
\subsection{Controllable image translation}
\noindent\textbf{Real image projection}\quad To edit an image in the latent space, we first need to project the image into the latent space. What matters here is how quickly and accurately the image can be reconstructed. We measured the runtime and LPIPS \cite{zhang2018unreasonable} between the input and reconstructed images. As shown in Table~\ref{tab:projection}, our model can embed an image into the latent space faster and more accurately than other real-time image editing methods.

\noindent\textbf{Style interpolation}\quad With the proposed method, it is possible to control only the style of the image as desired. In Fig.~\ref{fig:control} (a), we first projected images into content and style space, then interpolated style code with randomly selected prototypes. The results show that the proposed approach is suitable for controlling the results of synthesized images.
\tabproj

\noindent\textbf{Content transplantation}\quad Although we did not specifically target content transplantation, the proposed method supports this application. We achieved this by copying the content code from another content code. After manipulating the content code, we synthesized the image using a style code of the source image. As shown in Fig.~\ref{fig:control} (b), our model shows qualitatively similar results to the StyleMapGAN, which specifically targeting the local editing. Since our model separated the content and style, it is also possible to transplant only the content (\ie, a big smile) without changing the style (\ie, a beard) (bottom).

\noindent\textbf{Local image translation}\quad Fig.~\ref{fig:control} (c) shows the results of local image translation. The first two rows are the result of using vertically split masks. The red box in the bottom row indicates the mask for reference 1. The proposed method can synthesize the content using multiple styles.

\figsearch
\tabcombine
\subsection{Analysis}
\noindent\textbf{Effect of the style-aware discriminator}\quad We trained the model with a separated discriminator and style encoder to analyze the effect of integrating the discriminator and style encoder. The difference is that we used the hard-assigned prototypes as pseudo-label for the multi-task discriminator. To evaluate the alignment between learned style representation and domain labels, we measured the k-NN accuracy used for self-supervised learning \cite{he2020momentum,caron2021emerging}. In Table~\ref{tab:combine}, \texttt{separated} achieved significantly lower k-NN accuracy, and failed to relfect the style of the target images (high mFID). See Appendix ~\ref{appx:c1} for a further discussion.

\figablation
\noindent\textbf{Effect of data augmentation}\quad We employed random resized crop, rotation, and scale for augmentation, along with random erasing for facial datasets (\eg, CelebA-HQ, FFHQ).
Among them, we analyzed the effect of color distortion and cutout, which are major differences compared with other methods \cite{baek2021rethinking,lee2021contrastive}. As shown in Fig.~\ref{fig:search}, different augmentation choice leads to different style space. This result further leads to incorrect or unwanted synthesis results (Fig.~\ref{fig:ablation}). For example, when the color distortion is used, the style space ignores the color. On the other hand, if the cutout is not applied in the human face domain, learned style space failed to capture the attribute information such as gender.

\noindent\textbf{Speed and memory}\quad Table~\ref{tab:memory} shows the trainable parameter counts and the training time of each method. The proposed approach is more efficient and faster than conventional I2I translation methods because it requires one less module for training and has fewer modules than SwapAE, which uses two discriminators. Nevertheless, the proposed method achieved comparable or better performance, which shows the efficiency of our method.
\tabmemory
\section{Discussion and limitation}
\label{sec:discuss}
In this study, we proposed a \textit{style-aware discriminator}, which learns a style space in a self-supervised manner and guides the generator. Here, we discuss reasons why the proposed approach can be successfully trained. First, representation learning using human-defined labels cannot be a representation for style space. In contrast, the proposed method learns latent space specifically designed for style. Second, in the existing I2I translation, both the generator and the style encoder are updated together by the signal from the discriminator. In this case, the separation between content and style is ambiguous. Conversely, the proposed model can have a separate content space with the style encoder being updated completely separately from the generator, which results in better disentanglement. Finally, a style-aware discriminator can provide a better signal to the generator since it has a better understanding of the style space.

Yet still, the proposed method cannot preserve the face identity of the source image, unlike \cite{choi2020stargan,liu2021smoothing}. One can therefore consider using a pre-trained network for identity or landmark following previous works \cite{choi2020stargan,patashnik2021styleclip}. However, preserving the identity may increase risks of misuse or abuse. Therefore, we did not force the proposed method to preserve the facial identity of a source image. Though, preserving the facial identity without using additional information (\eg, face landmark or id) will be a valuable future work.

{\small
\noindent\textbf{Acknowledgements}\quad This work was supported by Institute of Information \& communications Technology Planning \& Evaluation (IITP) grant funded by the Korea government (MSIT) (No.B0101-15-0266, Development of High Performance Visual BigData Discovery Platform for Large-Scale Realtime Data Analysis) and (No.2017-0-00897, Development of Object Detection and Recognition for Intelligent Vehicles)
}

{\small
\bibliographystyle{ieee_fullname}
\bibliography{egbib}
}

\clearpage
\appendix
\section{Implementation}
\label{appx:a}
\subsection{Architecture}
\label{appx:a1}
The overall architecture of our method follows StarGAN v2 \cite{choi2020stargan}. We normalized the output content code in each pixel to the unit length following Park \etal \cite{park2020swapping}. When using the StyleGAN2-based generator, we replaced the instance normalization of the content encoder with pixel normalization \cite{karras2018progressive}. We did not use an equalized learning rate \cite{karras2020analyzing}.

The style-aware discriminator consists of $M = 0.25 * \log_2(\textrm{resolution})$ residual blocks followed by an average pooling. The style head and the discrimination head are two-layer MLPs. We used the same discriminator for the StyleGAN2-based and AdaIN-based models. We set dimension of prototypes to 256. We set $K$ to 32 for AFHQ, 64 for CelebA-HQ, and 128 for LSUN churches and FFHQ.

\subsection{Augmentation}
\label{appx:a2}
\noindent\textbf{Geometric transform}\quad We used the \texttt{RandomRation} and \texttt{RandomScaleAdjustment} augmentations. We applied reflection padding to avoid empty areas in the image before applying the geometric transform. We chose the rotation angle to be between -30 and 30 and the scale parameter between 0.8 and 1.2. Each transform was applied with a probability of 0.8.

\noindent\textbf{Cutout}\quad The style of the human face domain is integral to characteristics other than color and texture, including gender, expression, and accessories. We can read such information (\ie, the style) from an image even when part of a human face is occluded. Accordingly, we employed cutout augmentation. In practice, we used the \texttt{RandomErasing} method from the \texttt{torchvision} library with the following probability and scale parameters: \texttt{p=0.8} and \texttt{scale=(0.1, 0.33)}.

\noindent\textbf{Color distortion}\quad We observed that when the variation of the dataset is significant (\eg, FFHQ) or when the batch size is small, it was not possible to manipulate short hair into long hair. In that case, we employed weak color jittering. More specifically, we applied the \texttt{ColorJitter} method with the following parameters with a probability of 0.8: \texttt{brightness=0.2}, \texttt{contrast=0.2}, \texttt{saturation=0.2}, \texttt{hue=0.01}. Note that, we applied this augmentation only with the CelebA-HQ dataset using AdaIN and the FFHQ experiments.

\subsection{Style code sampling}
\label{appx:a3}
We sampled the style code from the dataset $\mathcal{X}$ with a probability $p$. Otherwise, we sampled from the prototypes. When sampling from a dataset, we used a randomly shuffled minibatch $\textbf{x}^{\prime}$ to create a style code $\tilde{\textbf{z}_s} = f_s(\textbf{x}^{\prime})$. In the case of sampling from the prototypes, we used the following pseudocode. In practice, we set $p$ to 0.8 except in the case of for longer training (25 M), where we used 0.5.
\begin{lstlisting}[language=Python]
# C: prototypes (K x D)
# N: batch size
# K: number of prototypes
# D: prototype dimension

@torch.no_grad()
def sample_from_prototypes(C, N, eps=0.01):
    K, D = C.shape

    samples = C[torch.randint(0, K, (N,))]
    if torch.rand(1) < 0.5:  # perturbation
        eps = eps * torch.randn_like(samples)
        samples = samples + eps
    else:  # interpolation
        targets = C[torch.randint(0, K, (N,))]
        t = torch.rand((N, 1))
        samples = torch.lerp(samples, targets, t)
    return F.normalize(samples, p=2, dim=1)
\end{lstlisting}

\subsection{Training details}
\label{appx:a4}
In every iterations, we sampled a minibatch $\textbf{x}$ of $N$ images from the dataset. To calculate the \textit{swapped prediction loss}, we created two different views $\textbf{x}_1 = \mathcal{T}_1(\textbf{x}), \textbf{x}_2 = \mathcal{T}_2(\textbf{x})$, where $\mathcal{T}$ is an augmentation. We reused the $\textbf{x}_1$ as the input of the generator. We obtained style codes by sampling the prototype with probability $p$ or encoding reference images $\textbf{x}^{\prime} = \textrm{\texttt{shuffle}}(\textbf{x}_1)$ with probability ($1-p$). In practice, we usually set $p$ as 0.8, but 0.5 when training is long enough (longer than 5 M). When sampling from the prototype, the first two of Eq. 2 was selected uniformly. The adversarial loss for updating the discriminator $D$ was calculated for $G(\textbf{x}_1, \textbf{s})$, and the adversarial loss for updating the generator $G$ was calculated for $G(\textbf{x}_1, \textbf{x})$ and the reconstructed image.

We applied the lazy R1 regularization following \cite{karras2020analyzing}. To stabilize the SwAV training, we adopted training details from the original paper \cite{caron2020unsupervised}. In more detail, we fixed the prototype for the first 500 iterations and used the queue after the 20,000th iteration if $K < N$. We linearly ramped up learning rate for the first 3000 iterations.
	
We initialized all of the networks using Kaiming initialization \cite{he2015delving}. Following Choi \etal \cite{choi2020stargan}, we used ADAM \cite{kingma2015adam} with a learning rate of 0.0001, $\beta_1 = 0.0$ and $\beta_2 = 0.99$. We scaled the learning rate of the mapping network by 0.01, similar to previous studies \cite{karras2019style,choi2020stargan}. By default, we used a batch size of 16 for the AdaIN-based model and 32 for the StyleGAN2-based model. We used a larger batch size (64) and longer training (25 M) for the FFHQ and LSUN churches datasets. We observed that the performance improves as the batch size and the number of training images increase.

\section{Additional results}
\label{appx:b}
\subsection{Quantitative results for the unlabeled datasets}
\tabunlabel
\tablerp
We measured the quality of the latent-guided and reference-guided synthesis on the unlabeled datasets in Table~\ref{tab:unlabel}. The proposed method significantly outperforms the Swapping Autoencoder \cite{park2020swapping} on the LSUN churches validation set. For reference, we also report the results of unconditionally generated StyleGAN2 images. Even though the proposed method is inferior to unconditional GANs (\ie, StyleGAN2 \cite{karras2020analyzing}), note that unconditional GANs are unsuitable for image editing \cite{park2020swapping}.

\subsection{Quality of the style interpolation}
\figlerp
To evaluate the quantitative results of the style interpolation, we calculated FID between the training set and images synthesized using interpolated styles ($\textrm{FID}_{lerp}$). We sampled images from two different domains and generated ten style codes by interpolating their corresponding style code. Then, we synthesized ten images using those style codes (we used the first sample as a source image). We created 30,000 fake images for the AFHQ and a total of 20,000 fake images for CelebA-HQ. As shown in Table~\ref{tab:lerp}, the proposed method outperforms the supervised approaches \cite{choi2020stargan,liu2021smoothing} in terms of FID. Fig.~\ref{fig:lerp} shows the qualitative comparison between the proposed model and baselines. The proposed approach was the only model that produced smooth interpolation results while maintaining the content such as backgrounds.

\subsection{Additional qualitative results}
Here, we include qualitative results for various datasets. Fig.~\ref{fig:afhqv2} shows the results of the model trained at 512$\times$512 resolution on the AFHQ v2 dataset. Fig.~\ref{fig:ffhq} and~\ref{fig:church} show the reference-guided image synthesis results on unlabeled datasets (FFHQ and LSUN churches). Fig.~\ref{fig:flower} shows the reference-guided image synthesis results for the Oxford-102 dataset. Finally, we visualize all prototypes learned with the AFHQ and CelebA-HQ datasets in Fig.~\ref{fig:proto}.

\section{Additional analyses}
\label{appx:c}
\subsection{Effect of the style-aware discriminator}
\label{appx:c1}
\figsep
The low k-NN metric of the \texttt{separated} method implies that the style space is not highly correlated with the species. This is further supported by the qualitative results. As shown in Fig.~\ref{fig:sep}, the \texttt{separated} method learns to translate the tone of the image rather than desired style (\ie, the species), which explains the very high mFID\footnote{In the AFHQ dataset, the models that cannot change species result in high mFID, since the FID between different species can be rather large. For example, the FID between a real cat and real dog is 170.4.}.

\subsection{Ablation based on the number of prototypes}
\tababalk
In Table~\ref{tab:abalk}, we evaluate the effect of the number of prototypes ($K$) on the proposed method. We trained the AdaIN-based model with varying $K$ using the AFHQ dataset. We observed that the appropriate number of prototypes was critical to the synthesis quality. However, even when the value of $K$ was large, the mFID value did not deviate from a certain range. We did not conduct experiments to determine the optimal value of $K$ for the other datasets; instead, we set the value of k based on the number of images in the dataset.

\figafhq
\figffhq
\figchurch
\figflower
\figproto

\end{document}